\documentclass[letterpaper]{article} 
\usepackage{aaai24}  
\usepackage{times}  
\usepackage{helvet}  
\usepackage{courier}  
\usepackage[hyphens]{url}  
\usepackage{graphicx} 
\usepackage{natbib}  
\usepackage{caption} 
\usepackage{algorithm}
\usepackage{algorithmic}
\usepackage{blindtext}
\usepackage{graphicx}
\usepackage{bm}
\usepackage{amssymb}
\usepackage{amsmath} 
\newtheorem{definition}{Definition}
\newtheorem{proposition}{Proposition}
\newtheorem{remark}{Remark}
\newtheorem{corollary}{Corollary}
\newtheorem{theorem}{Theorem}

\newtheorem{proof}{Proof}

\usepackage{csquotes}
\usepackage{color}

\usepackage{newfloat}
\usepackage{listings}
\DeclareCaptionStyle{ruled}{labelfont=normalfont,labelsep=colon,strut=off} 
\floatstyle{ruled}
\newfloat{listing}{tb}{lst}{}
\floatname{listing}{Listing}
\title{Exploring Gradient Explosion in Generative Adversarial Imitation Learning: \\A Probabilistic Perspective}
\author {
    Wanying Wang,\textsuperscript{\rm 1,}\equalcontrib\hspace{0.001cm}
    Yichen Zhu,\textsuperscript{\rm 2,}\equalcontrib\hspace{0.001cm}
    Yirui Zhou,\textsuperscript{\rm 1}
    Chaomin Shen,\textsuperscript{\rm 3}\\
    Jian Tang,\textsuperscript{\rm 2}
    Zhiyuan Xu,\textsuperscript{\rm 2}
    Yaxin Peng,\textsuperscript{\rm 1,$\dagger$}
    Yangchun Zhang\textsuperscript{\rm 1,$\dagger$}
}
\affiliations {
    \textsuperscript{\rm 1}Department of Mathematics, College of Sciences, Shanghai University\\
    \textsuperscript{\rm 2}Midea Group\\
    \textsuperscript{\rm 3}School of Computer Science, East China Normal University\\
    \{wywang, zyr050798, yaxin.peng, zycstatis\}@shu.edu.cn,
    cmshen@cs.ecnu.edu.cn,\\ \{zhuyc25, xuzy70, tangjian22\}@midea.com
}

\begin{document}
\maketitle

\begin{abstract}
Generative Adversarial Imitation Learning (GAIL) stands as a cornerstone approach in imitation learning. This paper investigates the gradient explosion in two types of GAIL: GAIL with deterministic policy (DE-GAIL) and GAIL with stochastic policy (ST-GAIL). We begin with the observation that the training can be highly unstable for DE-GAIL at the beginning of the training phase and end up divergence. Conversely, the ST-GAIL training trajectory remains consistent, reliably converging. To shed light on these disparities, we provide an explanation from a theoretical perspective. By establishing a probabilistic lower bound for GAIL, we demonstrate that gradient explosion is an inevitable outcome for DE-GAIL due to occasionally large expert-imitator policy disparity, whereas ST-GAIL does not have the issue with it. To substantiate our assertion, we illustrate how modifications in the reward function can mitigate the gradient explosion challenge. Finally, we propose CREDO, a simple yet effective strategy that clips the reward function during the training phase, allowing the GAIL to enjoy high data efficiency and stable trainability. 
\end{abstract}

\section{Introduction}
Imitation learning trains a policy directly from expert demonstrations without reward signals \cite{ng2000algorithms,syed2007a,ho2016generative}. It has been broadly studied under the twin umbrellas of behavioral cloning~\cite{pomerleau1991efficient} and inverse reinforcement learning (IRL)~\cite{ziebart2008maximum}. Generative adversarial imitation learning (GAIL)~\cite{ho2016generative}, established by the policy training of trust region policy optimization~\cite{schulman2015trust}, plugs the inspiration of generative adversarial networks~\cite{goodfellow2014generative} into the maximum entropy IRL. The discriminator in GAIL aims to distinguish whether a state-action pair comes from the expert demonstration or is generated by the agent. Meanwhile, the learned policy generates interaction data to confuse the discriminator. GAIL is promising for many real-world scenarios where designing reward functions to learn the optimal control policies requires significant effort. It has made remarkable achievements in physical-world tasks, i.e., robot manipulation \cite{jabri2021robot}, mobile robot navigating \cite{tai2018socially}, commodities search \cite{shi2019virtual} and endovascular catheterization \cite{chi2020collaborative}. 

The GAIL can be bifurcated into two genres: stochastic policy algorithms and deterministic policy algorithms, namely DE-GAIL~\cite{kostrikov2019discriminator,zuo2020deterministic} and ST-GAIL~\cite{ho2016generative, zhou2022generalization}. The ST-GAIL with stochastic policy guarantees global convergence in high-dimensional environments, outperforming traditional Inverse Reinforcement Learning (IRL) methods~\cite{ng2000algorithms,ziebart2008maximum,boularias2011relative}. Nevertheless, its application in real-world scenarios is limited due to low sample efficiency and excessive training times~\cite{zuo2020deterministic}. On the contrary, DE-GAIL has become a preferred approach due to its exceptional data efficiency. Typically, it outpaces GAIL with stochastic policy by a factor of more than ten, significantly accelerating the learning process.

However, while the DE-GAIL is much more data-efficient than ST-GAIL, it is not flawless: we observe a significant likelihood of generating near-zero rewards from the very beginning of the training stage. To elucidate this, we carried out experiments on three environments in Mujuco with multiple DE-GAIL and ST-GAIL algorithms. Among 11 experiments under uniform training settings, we observed DE-GAIL method frequently failed during the training phase. The average divergence rate is over 36\% (Details can be found in Section \ref{sec:2}).

\begin{figure*}[ht]
\centerline{
\includegraphics[width=0.85\linewidth]{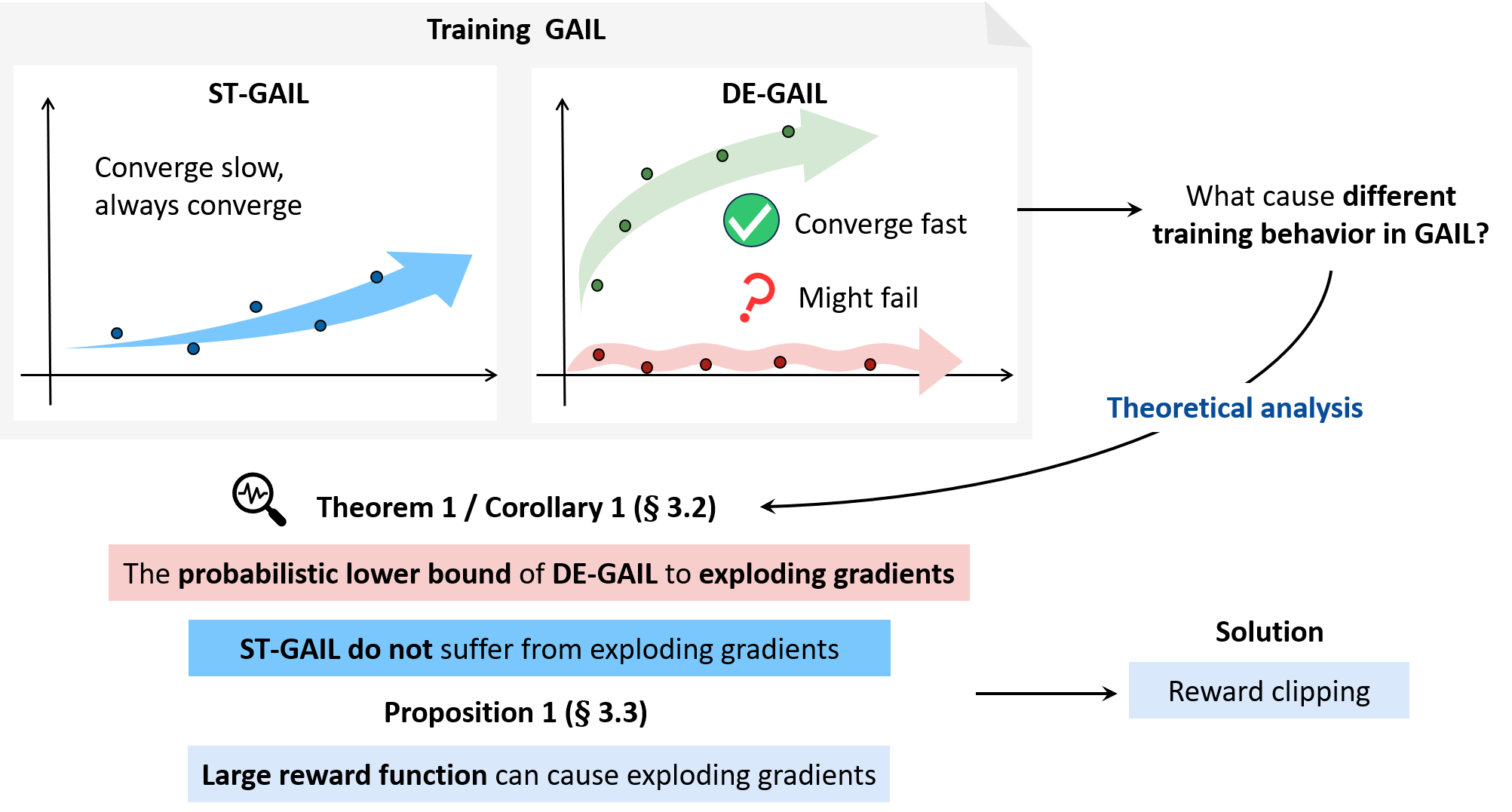}}
\caption{The overview of our analysis. Theorem 1 and Corollary 1 develop the probabilistic lower bound of DE-GAIL to quantify the gradient explosion. Proposition 1 connects the gradient explosion with the reward function. We further present a reward clipping technique to relieve the gradient explosion issue in DE-GAIL.}
\label{fig:architecture}
\end{figure*}

Why does DE-GAIL have such a high probability of diverging? To shed light on this question, we prove a probabilistic lower bound that describes the gradient explosion in DE-GAIL. Our proof is built upon the policy disparity between expert demonstration and imitative action. In short, if the agent fails to mimic the expert's action, the expert has a large reward; then, the gradient could explode during training. We verify our conclusion by showing a simple manipulation of the reward function, i.e., switching to adversarial inverse reinforcement learning (AIRL)~\cite{fu2018learning}, which can alleviate the gradient explosion issue in GAIL. Nevertheless, since the DE-GAIL is much more data-efficient than ST-GAIL, we seek to resolve this issue by proposing CREDO. This reward clipping technique is both empirically effective and theoretically sound. 

Our overall framework for analyzing gradient explosion in GAIL is shown in Fig.~\ref{fig:architecture}. In summary, our contributions are the follows:
\begin{itemize}
\item We conduct a comprehensive empirical study to show the fact that DE-GAIL is training unstable yet converges fast. On the other hand, the ST-GAIL is data inefficient yet ensures convergence. 
\item We develop a series of theoretical proofs to support our observation and conclude that reward function is the cause of the gradient explosion in DE-GAIL. 
\item We present a simple technique called CREDO which clips the reward function during training to relieve the gradient explosion problem in DE-GAIL. 
\end{itemize}

\section{Evidence of Gradient Explosion in GAIL}\label{sec:2}
In this section, we perform a comprehensive study to examine the gradient explosion issue in GAIL. We reproduced three environments in Mujoco \cite{todorov2012mujoco}, Hopper-v2, HalfCheetah-v2, and Walker2d-v2, following the setup in two-stage stochastic gradient (TSSG)~\cite{zhou2022generalization}. The expert trajectories were generated via the soft actor-critic (SAC) agent \cite{haarnoja2018soft2}. The expert demonstration has one million data points with a standard deviation of 0.01. We repeat our experiments 11 times across all environments, maintaining a consistent training setting, except for the number of random seeds. 

Regarding the network architecture, we employ two-layer networks designed to approximate the kernel function~\cite{arora2019exact} to train GAIL. The reward function is defined as $r(s,a)=-\log(1-D(s,a))$, which is referred to as Probability Logarithm Reward (PLR). Our evaluations encompass three variants of the DE-GAIL and two ST-GAIL methods. The DE-GAIL methods include deep deterministic policy gradient~\cite{lillicrap2015continuous} (DDPG-GAIL), twin delayed deep deterministic policy gradient~\cite{fujimoto2018addressing} (TD3-GAIL), and softmax deep double deterministic policy gradients~\cite{pan2020softmax} (SD3-GAIL). The first two are recognized and widely adopted DE-GAIL algorithms, whereas SD3-GAIL represents a more recent and refined approach. For ST-GAIL, we utilize proximal policy optimization~\cite{schulman2017proximal} (PPO)-GAIL ~\cite{chen2020on} and TSSG~\cite{zhou2022generalization}, a method that integrate SAC into GAIL. 

\begin{figure*}[t]
\centerline{
\includegraphics[width=\textwidth]{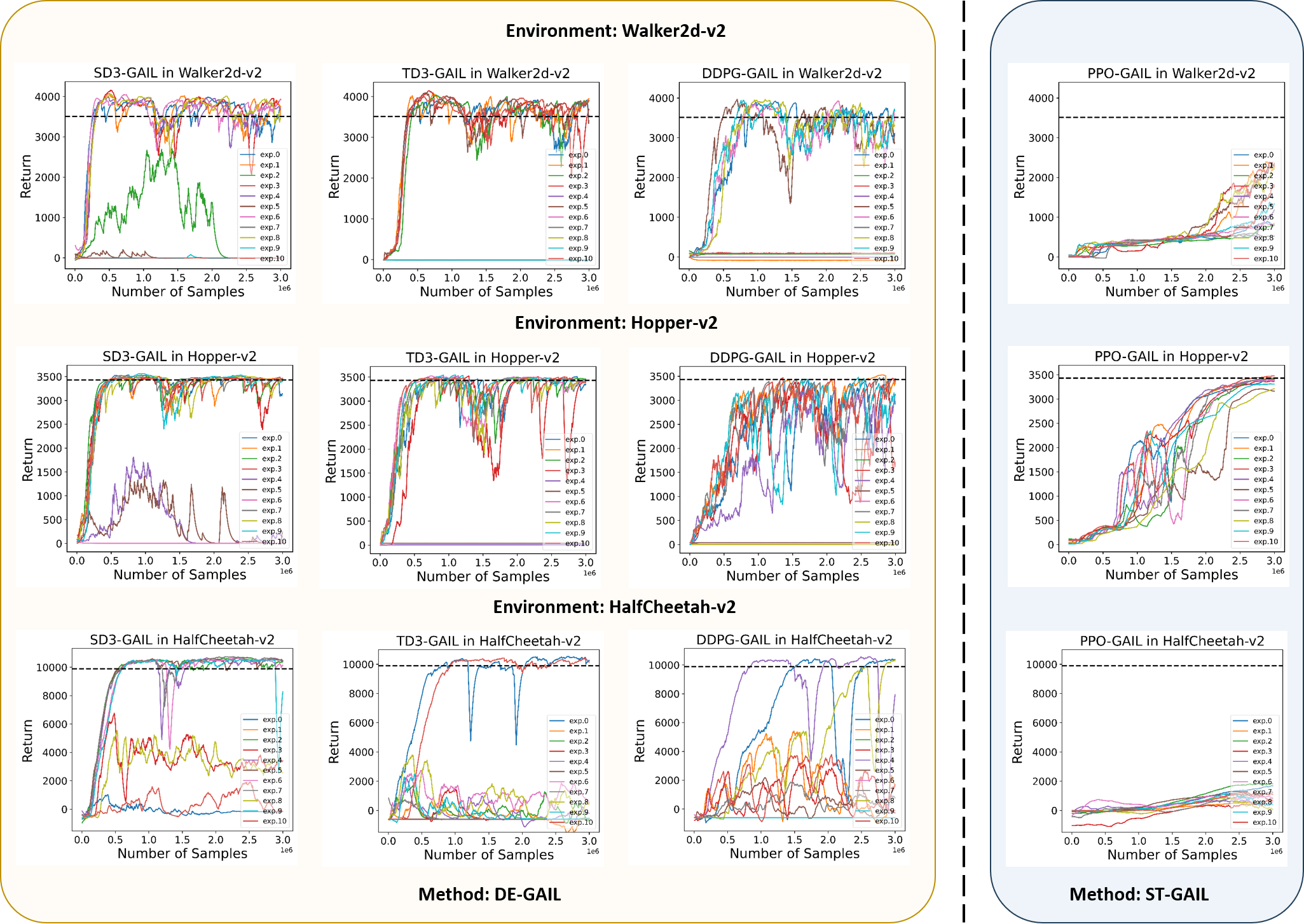}
}
\caption{We conduct 11 experiments spread across three environments, with three DE-GAIL and ST-GAIL methods. We observe a clear tendency for DE-GAIL algorithms to struggle to reach convergence during the training phase in multiple experiments. In comparison, the training procedure utilizing the ST-GAIL method showcased significantly higher stability.}
\label{fig:4algo_3env_11seed}
\end{figure*}

As shown in Fig.~\ref{fig:4algo_3env_11seed}, we observed that all three DE-GAIL algorithms could potentially fail during training, irrespective of how advanced they are. This pattern is consistently seen across all three tasks. It illustrates that the training of DE-GAIL methods can be particularly unstable at the initial stages, reaching a point from which recovery becomes impossible as training progresses. This behavior sharply contrasts with the training curve of successful experiments, which, on the other hand, converge quickly and yield high returns, thus highlighting the solid data-efficiency characteristic of the DE-GAIL approach. Turning our attention to ST-GAIL, we noted that even though its convergence speed lags behind DE-GAIL, all experiments exhibited consistent and successful convergence.

In summary, our observations reveal the following insights:
\begin{itemize}
    \item The initial phase of DE-GAIL training can be remarkably unstable. However, once convergence is attained, it is often swift and results in higher return values, potentially leading to an elevated success rate.
    \item Conversely, ST-GAIL exhibits stability during the initial training process, yet its convergence is approximately ten times slower and it tends to achieve lower return values compared to advanced DE-GAIL algorithms.
\end{itemize}
These phenomena prompt us to delve into the root causes underpinning their differences. In the subsequent section, we offer a theoretical framework to support and deepen our understanding of these observations.

\section{Gradients Explosion in GAIL: A Probability View\label{section_proof}}
In this section, we will first introduce the necessary background information and notation for the forthcoming proof. Then, we provide a detailed analysis from a theoretical standpoint along with empirical evidence to unveil the mystery of gradient explosion in GAIL. 

\subsection{Background and Annotation}
\subsubsection{Markov Decision Process}
A discounted Markov Decision Process (MDP) in the conventional Reinforcement Learning (RL) context is defined by a quintuple $(\mathcal{S},\mathcal{A},r,p_{M},\gamma)$. Here, $\mathcal{S}$ and $\mathcal{A}$ represent the finite state space and action space respectively. The reward function, $r(s,a): \mathcal{S}\times \mathcal{A}\rightarrow \mathbb{R}$, denotes the reward obtained from executing action $a\in \mathcal{A}$ in state $s\in \mathcal{S}$. The transition distribution is represented by $p_{M}(s'|s,a): \mathcal{S}\times \mathcal{A}\times \mathcal{S}\rightarrow [0,1]$, and $\gamma$ is the discount factor.

A stochastic policy, denoted as $\pi(a|s)$, can be characterized as a probability function mapping a state $s\in \mathcal{S}$ to a distribution of action $a\in \mathcal{A}$, expressed formally as $\mathcal{S}\times \mathcal{A}\rightarrow [0,1]$. In contrast, a deterministic policy, $\pi(s)$, is defined as a direct mapping from a state $s\in \mathcal{S}$ to a corresponding action $a\in \mathcal{A}$, formally written as $\mathcal{S}\rightarrow \mathcal{A}$.

The primary objective of Reinforcement Learning (RL) is to maximize the expected reward-to-go, represented as $\eta(\pi)=\mathbb{E}_{\pi}\left[\sum_{t=0}^{\infty} \gamma^{t} r\left(s_{t}, a_{t}\right) | s_{0}, a_{0} \right]$. Induced by a policy $\pi$, \emph{the discounted stationary state distribution} is defined as $d^{\pi}(s)=(1-\gamma)\sum_{t=0}^{\infty}\gamma ^{t}{\rm Pr}(s_{t}=s;\pi)$. Similarly, \emph{the discounted stationary state-action distribution} is given by $\rho^{\pi}(s,a)=(1-\gamma)\sum_{t=0}^{\infty}{\gamma ^{t}{\rm Pr}(s_{t}=s,a_{t}=a;\pi)}$. This distribution measures the cumulative \enquote{frequency} with which a state-action pair is visited under the policy $\pi$.

Let $\mathrm{P} \in \mathbb{R}_{+}^{|\mathcal{S}||\mathcal{A}| \times|\mathcal{S}|}$ denote the transition matrix where $\mathrm{P}_{sa, s'}= p_{M}(s' \mid s, a)$, 
$\boldsymbol{\pi} \in \mathbb{R}_{+}^{|\mathcal{S}||\mathcal{A}| \times 1}$ denote the policy matrix where $\left(\boldsymbol{\pi}\right)_{s a}=\pi(a|s)$, and $\boldsymbol{\pi}_{s_i}\in \mathbb{R}_{+}^{|\mathcal{A}| \times 1}$ the policy for the state $s_i$. Define the expanded matrix of $\boldsymbol{\pi}$ as
\begin{align}
\Pi=\left[\begin{array}{ccc}
\boldsymbol{\pi}_{s_1}^{\top} & & \mathbf{0} \\
& \ddots & \\
\mathbf{0} & & \boldsymbol{\pi}_{s_{|\mathcal{S}|}}^{\top}
\end{array}\right] \in \mathbb{R}_{+}^{|\mathcal{S}| \times|\mathcal{S}||\mathcal{A}|}.
\end{align}
We use $\boldsymbol{\pi}_h$ to denote a parameterized policy matrix with parameters $h$, then the policy Jacobian matrix $\mathrm{H}_h \in \mathbb{R}^{1 \times |\mathcal{S}||\mathcal{A}|}$ is $\left(\mathrm{H}_h\right)_{ s a}=\nabla_{h}\left(\boldsymbol{\pi}_h\right)_{s a}=\nabla_{h}\pi_{h}(a|s)$. The state-action distribution matrix and the state distribution matrix are $\bm{\rho}^{h} \in \mathbb{R}_{+}^{|\mathcal{S} \| \mathcal{A}| \times 1}$ and $\mathbf{d}^{h} \in \mathbb{R}_{+}^{|\mathcal{S} | \times 1}$, respectively, where $(\bm{\rho}^{h})_{sa}=\rho^{\pi_h}(s, a)$ and $(\bm{d}^{h})_{s}=d^{\pi_h}(s)$. More precisely, $\mathbf{d}^{h}=T \bm{\rho}^{h}$, here $T$ is the marginalization matrix
\begin{align}
T=\left[\begin{array}{ccc}
\mathbf{1}_{|\mathcal{A}|}^{\top} & & \mathbf{0} \\
& \ddots & \\
\mathbf{0} & & \mathbf{1}_{|\mathcal{A}|}^{\top}
\end{array}\right] \in \mathbb{R}^{|\mathcal{S}| \times|\mathcal{S}||\mathcal{A}|},
\end{align}
where $\mathbf{1}$ is the vector of all ones and the
subscript represents its dimensionality.

\subsubsection{Generative Adversarial Imitation Learning (GAIL)}
The description of GAIL can be found in Supplementary Material D. Here, we define the discriminator $D(s,a)$, the imitative policy $\pi$, and the expert policy $\pi_{\rm E}$. Given a state, the discriminator quantifies the distributional disparity between the expert's and imitative policies. This disparity can be interpreted as a reward for the agent. Consequently, the optimization problem for GAIL can be formulated as follows:
\begin{align}	
&\mathop{\min}_{\pi}\mathop{\max}_{D\in(0,1)^{\mathcal{S}\times \mathcal{A}}}~\mathbb{E}_{(s,a)\sim \rho^{\pi_{\rm E}}}[\log(D(s,a))] \notag\\
&~~~~~~~~~~~~~~~~~~~~~~~~~~~~~~~~~~+\mathbb{E}_{(s,a)\sim \rho^{\pi}}[\log(1-D(s,a))],
\label{gail}
\end{align}
where the policy $\pi$ mimics the expert policy via the reward function $r(s,a)=-\log(1-D(s,a))$. When the discriminator reaches its optimum,
\begin{align}
D^{\ast}(s,a)=\rho^{\pi_{\rm E}}(s,a)/ (\rho^{\pi_{\rm E}}(s,a)+\rho^{\pi}(s,a)),
\label{D_optimum}
\end{align}
the optimization objective for the learned policy is formalized as minimizing the discrepancy in the state-action distribution between the imitated policy and the expert policy. This discrepancy is quantified using the Jensen-Shannon (JS) divergence, 
\begin{align}
&\mathop{\min}_{\pi}{D_{\rm JS}(\rho^{\pi}(s,a),\rho^{\pi_{\rm E}}(s,a))}:=\frac{1}{2}D_{\rm KL}\left(\rho^{\pi},\frac{\rho^{\pi}+\rho^{\pi_{\rm E}}}{2}\right) \notag\\
&~~~~~~~~~~~~~~~~~~~~~~~~~~~+ \frac{1}{2}D_{\rm KL}\left(\rho^{\pi_{\rm E}},\frac{\rho^{\pi}+\rho^{\pi_{\rm E}}}{2}\right).
\end{align}

\subsection{Exploding Gradients in GAIL}
\label{3.2}
We employ multivariate Gaussian policy to approximate deterministic policy~\cite{paternain2020stochastic, lever2015modelling}, where the learned policy $\pi_{h}$ is defined as follow:
\begin{align}
\pi_{h}(a|s)=\frac{1}{\sqrt{\det(2\pi\bm{\Sigma})}}\exp{\frac{-(a-h(s))^{\top}\bm{\Sigma}^{-1}(a-h(s))}{2}},
\end{align}
The above equation is parameterized by deterministic functions $h:\mathcal{S}\rightarrow\mathcal{A}$ and covariance matrix $\bm{\Sigma}$. The function $h(\cdot)$ is an element of reproducing kernel Hilbert space $\mathcal{H}_{\kappa}$, $h(\cdot)=\sum_{i}\kappa(s_{i},\cdot)a_{i}\in \mathcal{H}_{\kappa}$, where $\kappa(s_{i}, s_{j})$ is the kernel function, $s_{i}\in \mathcal{S}$ and $a_{i}\in \mathcal{A}$. Note that $\pi_{h}(a|s)$ can be regarded as an approximation to the Dirac’s impulse via covariance matrix approaching zero, for instance, 
\begin{align}
\lim_{\bm{\Sigma} \rightarrow \bm{0}}\pi_{h}(a|s)=\delta(a-h(s)).
\label{dirac}
\end{align}
Eq. (\ref{dirac}) means that when the covariance $\bm{\Sigma} \rightarrow \bm{0}$, the stochastic policy $\pi_{h}(a|s)$ approaches the deterministic policy $h(s)$. Therefore, we can substitute $\pi$ with $\pi_{h}$ and rewrite the optimization problem of GAIL under $\pi_{h}$ is
\begin{align}
&\mathop{\min}_{\pi_{h}}\mathop{\max}_{D}~
\mathbb{E}_{(s,a)\sim \rho^{\pi_{\rm E}}}[\log(D(s,a))] \notag\\
&~~~~~~~~~~~~~~~~~~~~~~~~~~~~~~~~~+\mathbb{E}_{(s,a)\sim \rho^{\pi_{h}}}[\log(1-D(s,a))],
\end{align}
the optimal discriminator is
\begin{align}
D^{\ast}(s,a)=\rho^{\pi_{\rm E}}(s,a)/ (\rho^{\pi_{\rm E}}(s,a)+\rho^{\pi_{h}}(s,a)),
\label{D_ast}
\end{align}
and the policy optimization objective is
\begin{align}
\mathop{\min}_{\pi_{h}}{D_{\rm JS}(\rho^{\pi_{h}}(s,a),\rho^{\pi_{\rm E}}(s,a))}.
\end{align}
Before jumping into our main result, we need the following definition.
\begin{definition}
[Expert-Imitator Policy Disparity] Given the state $s_{t}$ at time $t$, 
$a_{t}$ and $h(s_t)$ are the actions induced by the expert policy and the imitated policy, respectively. If $\|h(s_{t})-a_{t}\|_2\geq C\|\bm{\Sigma}\|_{2}$ for any $C>0$, we say that there exist policy disparity between the expert and the imitator. Otherwise, the $(s_{t},h(s_{t}))$ perfectly matches the $(s_{t},a_{t})$. 
\end{definition}
Here, we utilize an event
\begin{align}
\Xi=\{(s_{t},h(s_{t})):\|h(s_{t})-a_{t}\|_2\geq C\|\bm{\Sigma}\|_{2}~\text{for any } C>0\}
\label{event}
\end{align}
to characterize the expert-imitator policy disparity. For convenience, we will use policy disparity to denote such behavior. Now we present the following theorem on the probability of exploding gradients in DE-GAIL. 
\begin{theorem}
Let $\pi_{h}(\cdot|s)$ be the Gaussian stochastic policy with mean $h(s)$ and covariance $\bm{\Sigma}$. When the discriminator achieves its optimum $D^{\ast}(s,a)$ in Eq. (\ref{D_ast}), the gradient estimator of the policy loss with respect to the policy’s parameter $h$ satisfies $\|\hat{\nabla}_{h}D_{\rm JS}(\rho^{\pi_{h}},\rho^{\pi_{\rm E}})\|_2\rightarrow \infty$ with a probability of at least $\Pr(\|\bm{\Sigma}^{-1}(a_{t}-h(s_{t}))\|_2\geq C~\text{for any } C>0)$ as $\bm{\Sigma}\rightarrow \bm{0}$, where 
\begin{align*}
\hat{\nabla}_{h}D_{\rm JS}(\rho^{\pi_{h}},\rho^{\pi_{\rm E}})
=&\frac{H_h \Delta\left(T^{\top} \mathbf{d}^{h}\right) (\mathrm{I}-\gamma \mathrm{P} \Pi_h)^{-1}\mathbf{e}_{s_t,a_t}}{2\rho^{\pi_{\rm E}}(s_t,a_t)}\cdot \notag\\
&~~~~~~~~~~~~\log\frac{2\rho^{\pi_{h}}(s_t,a_t)}{\rho^{\pi_{h}}(s_t,a_t)+\rho^{\pi_{\rm E}}(s_t,a_t)},
\end{align*}
and $\left(\mathrm{H}_h\right)_{ s a}
=\pi_{h}(a|s)\kappa(s,\cdot)\bm{\Sigma}^{-1}(a-h(s))$, $\Delta(\cdot)$ maps a vector to a diagonal matrix with its elements on the main diagonal, $\mathbf{e}_{s_t,a_t}=[0,\cdots,\mathop{1}_{s_t,a_t},\cdots,0]^{\top} \in \mathbb{R}^{|\mathcal{S}||\mathcal{A}|\times 1}$. 
\label{optimal_theorem}
\end{theorem}
\noindent \textbf{Proof~}
See Supplementary Material A.1. 

The result establishes the probability of exploding gradients in DE-GAIL. Due to the compatibility of norms, we have 
\begin{align}
&~~~~\Pr(\|\bm{\Sigma}^{-1}(a_{t}-h(s_{t}))\|_2\geq C~\text{for any } C>0) \notag\\
&\geq \Pr(\|a_{t}-h(s_{t})\|_2\geq C\|\bm{\Sigma}\|_2~\text{for any } C>0) \notag\\
&=\Pr(\Xi).
\end{align}
Therefore, the probability of policy disparity $\Pr(\Xi)$ constitutes a probabilistic lower bound of exploding gradients in DE-GAIL. Note that $\Pr(\Xi)$ is nontrivial as $\bm{\Sigma} \rightarrow \bm{0}$. 

\begin{remark}

Theorem \ref{optimal_theorem} implies that when the discriminator achieves its optimal state, DE-GAIL will suffer from exploding gradients with the probabilistic lower bound $\Pr(\Xi)>0$.

In contrast, for a Gaussian stochastic policy (fixed $\bm{\Sigma}$), we have that $\|\hat{\nabla}_{h}D_{\rm JS}(\rho^{\pi_{h}},\rho^{\pi_{\rm E}})\|_2$ is bounded referring to the proof strategy of Theorem \ref{optimal_theorem}. Thus, when the discriminator achieves its optimal state, the Gaussian stochastic policy in GAIL will not suffer from exploding gradients.
\end{remark}

Theorem \ref{optimal_theorem} indicates that when the discriminator attains its optimal state, the policy loss can encounter a gradient explosion issue. However, this represents an idealized scenario. In practical applications, the discriminator seldom reaches this optimum. Hence, we adapt our findings to a broader context using a \enquote{non-optimum} discriminator derived from data. Here, we name such \enquote{non-optimum} discriminators as imperfect discriminators and define them by the following: 
\begin{align}
\label{Tilde_D}
\Tilde{D}(s_{t},a_{t})&=\frac{(1+\epsilon_{1})\rho^{\pi_{\rm E}}(s_{t},a_{t})}{(1+\epsilon_{1})\rho^{\pi_{\rm E}}(s_{t},a_{t})+(1-\epsilon_{2})\rho^{\pi}(s_{t},a_{t})},  \notag\\
&\text{where}
\left \{
\begin{array}{l}
\epsilon_{1}> -1, \epsilon_{2}< 1\\
\epsilon_{1}< -1, \epsilon_{2}> 1 
\end{array}
\right. 
\end{align}

The explanation of the imperfect discriminator $\Tilde{D}(s_{t},a_{t})$ and its properties are as follows:
\begin{itemize}
\item $\epsilon_{1}$ and $\epsilon_{2}$ can be regarded as fluctuations in the optimal discriminator.
\item The imperfect discriminator generalize the fixed $D^{\ast}(s_{t},a_{t})$ to a ranges within $(0,1)$ stemmed from Eq. (\ref{gail}).
\item $\Tilde{D}(s_{t},a_{t})$ degenerates to 0 when $\epsilon_1=-1$ and degenerates to 1 when $\epsilon_2=1$.
\item $\Tilde{D}(s_{t},a_{t})$ reaches its optimum when $\epsilon_{1}$ and $\epsilon_{2}$ are 0.
\end{itemize}

We next state the exploding gradients on the imperfect discriminator $\Tilde{D}(s_{t}, a_{t})$.

\begin{corollary}
Let $\pi_{h}(\cdot|s)$ be the Gaussian stochastic policy with mean $h(s)$ and covariance $\bm{\Sigma}$. When the discriminator is in the format of Eq. (\ref{Tilde_D}), i.e., $\Tilde{D}(s,a)\in (0,1)$, the gradient estimator of the policy loss concerning the policy’s parameter $h$ satisfies 
\begin{align*}
\left\| \hat{\nabla}_{h}\left( \mathbb{E}_{\mathcal{D}^{\star}}[\log \tilde{D}(s,a) ]+\mathbb{E}_{\mathcal{D}}[\log(1-\tilde{D}(s,a))] \right) \right\|_{2}\rightarrow \infty
\end{align*}
with a probability of at least
\begin{align*}
\Pr(\|\bm{\Sigma}^{-1}(a_{t}-h(s_{t}))\|_2\geq C~\text{for any } C>0)
\end{align*}
as $\bm{\Sigma}\rightarrow \bm{0}$, where $\mathcal{D}^{\star}$ and $\mathcal{D}$ denote the expert demonstration and the replay buffer of $\pi_{h}$ respectively, 
\begin{align*}
&~~~~\hat{\nabla}_{h}\left( \mathbb{E}_{\mathcal{D}^{\star}}[\log(\tilde{D}(s,a))]+\mathbb{E}_{\mathcal{D}}[\log(1-\tilde{D}(s,a))] \right) \\
&=\frac{H_h \Delta\left(T^{\top} \mathbf{d}^{h}\right) (\mathrm{I}-\gamma \mathrm{P} \Pi_h)^{-1}\mathbf{e}_{s_t,a_t}}{\rho^{\pi_{\rm E}}(s_{t},a_{t})}\cdot\\
&~~~~~~~~~~~~\log \frac{(1-\epsilon_{2})\rho^{\pi_{h}}(s_t,a_t)}{(1+\epsilon_{1})\rho^{\pi_{\rm E}}(s_{t},a_{t})+(1-\epsilon_{2})\rho^{\pi_{h}}(s_t,a_t)} \\
&~~~~+\frac{(\epsilon_{1}+\epsilon_{2})\nabla_{h}\rho^{\pi_{h}}(s_t,a_t)}{(1+\epsilon_{1})\rho^{\pi_{\rm E}}(s_{t},a_{t})+(1-\epsilon_{2})\rho^{\pi_{h}}(s_{t},a_{t})},
\end{align*}
and $\left(\mathrm{H}_h\right)_{ s a}
=\pi_{h}(a|s)\kappa(s,\cdot)\bm{\Sigma}^{-1}(a-h(s))$, $\Delta(\cdot)$ maps a vector to a diagonal matrix with its elements on the main diagonal, $\mathbf{e}_{s_t,a_t}=[0,\cdots,\mathop{1}_{s_t,a_t},\cdots,0]^{\top} \in \mathbb{R}^{|\mathcal{S}||\mathcal{A}|\times 1}$. 
\label{any_D_theorem}
\end{corollary}

\noindent \textbf{Proof~} See Supplementary Material A.2.\\

Analogous to Theorem \ref{optimal_theorem}, Corollary \ref{any_D_theorem} suggests that when the discriminator adopts the perturbation form given by Eq. (\ref{Tilde_D}), DE-GAIL is susceptible to gradient explosions. Conversely, ST-GAIL remains unaffected by such explosions as long as the discriminator values lie within the interval $(0,1)$.

\begin{remark}
In the implementation of deterministic policy, it requires exploration that adds noise $\Sigma'$ to the output action. Concurrently, as the deterministic policy $h$ progressively updated, the covariance of Gaussian stochastic policy $\Sigma \rightarrow 0$. Notably, the stochastic factors are taken into consideration by using 
\begin{align}
\Xi_{1} =&\{(s_{t},h(s_{t})):\|h(s_{t})+\mathcal{N}(0,\Sigma')-a_{t}\|_{2}\geq C\|\Sigma\|_2 \notag\\
&\text{ for any } C>0\}
\end{align}
to characterize policy disparity in practice. During $\Sigma \rightarrow 0$, the proofs of Theorem \ref{optimal_theorem} and Corollary \ref{any_D_theorem} only depend on $\Sigma \rightarrow 0$, Theorem \ref{optimal_theorem} and Corollary \ref{any_D_theorem} both hold for $\Xi$ and $\Xi_{1}$ regardless of $\Sigma'$.
\end{remark}
In brief, our Theorem 1 and Corollary 1 hold in the practical setting. 
\begin{figure*}[ht]
\centering
\includegraphics[width=0.9\linewidth]{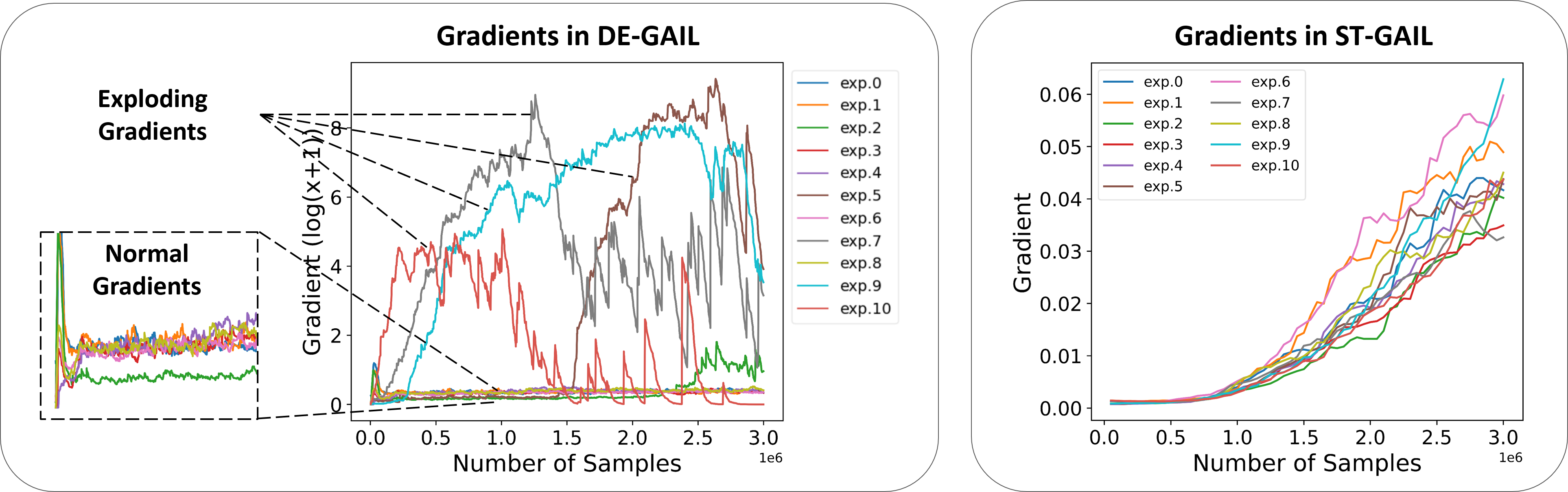}
\caption{The absolute gradients of SD3-GAIL and PPO-GAIL policy networks in Walker2d-v2. Each algorithm is repeatedly run 11 times. We observe that four experiments in SD3-GAIL display exploding gradients. Note that we use $log(x+1)$ to rescale the y-axis in the left figure.}
\label{gradients}
\end{figure*}

\noindent
\textbf{Empirical evidence.} Our experimental results support our theoretical analysis. Specifically, in Fig~\ref{gradients}, we record the absolute gradient in the training phase. Notably, four experiments in SD3-GAIL have gradient explosions. The total percent of failure cases is over 36\%.

\begin{figure}[ht]
\centerline{
\includegraphics[width=\linewidth]{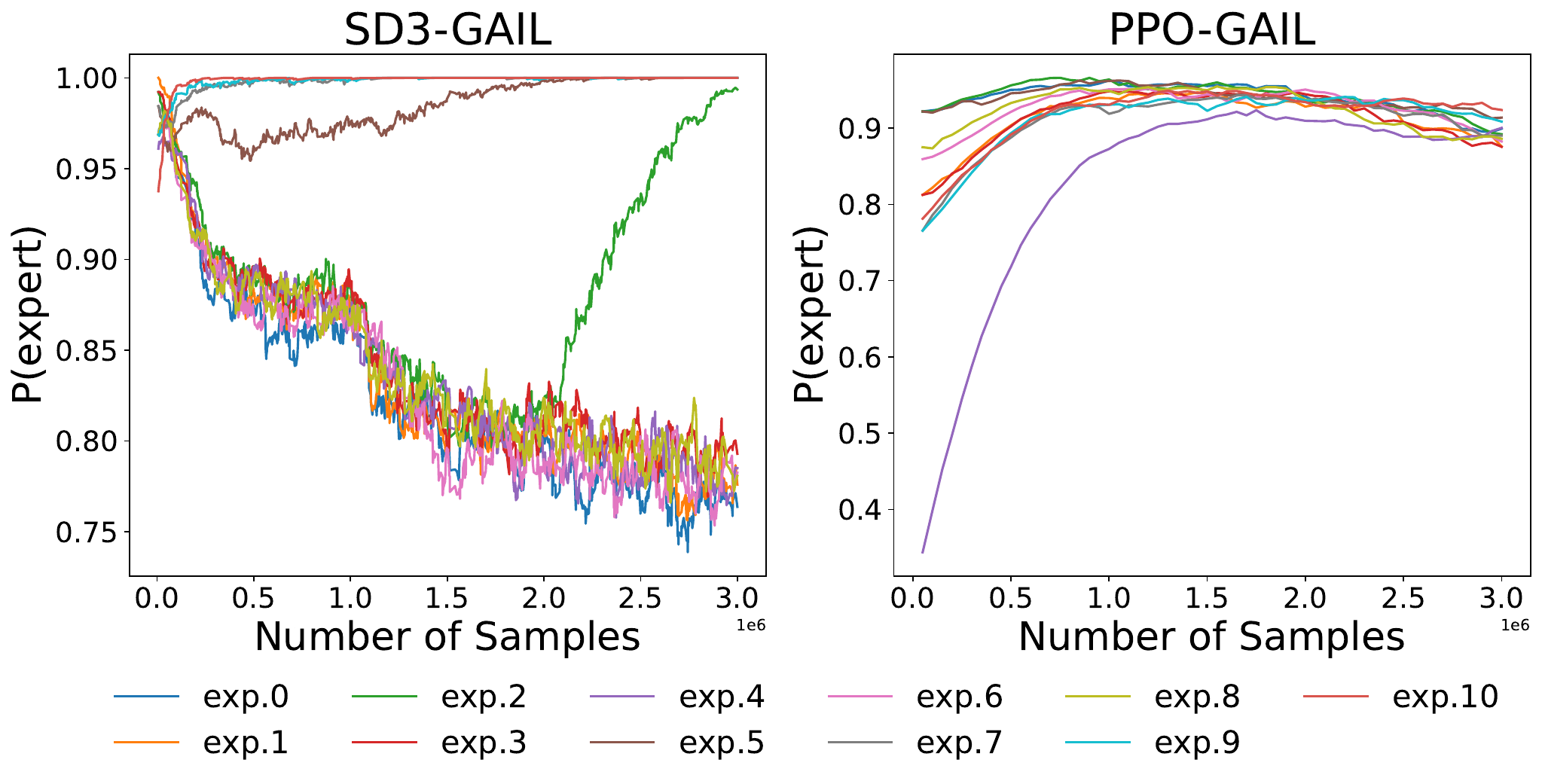}}
\caption{The y-axis denotes the probability of the imitator's policy being classified as the expert's policy, recorded as $P\text{(expert)}$. We observe that DE-GAIL can have $P\text{(expert)} \approx 1$, which leads to gradient explosion during training.}
\label{disc}
\end{figure}

In Fig.\ref{disc}, we document the probability $P\text{(expert)}$ of the expert's demonstration being classified to expert policy by the discriminator. It's crucial to understand that when $P\text{(expert)} \approx 1$, there's potential for gradient explosion. Our observations indicate that the $P\text{(expert)}$ of ST-GAIL never attains a value of one. On the contrary, in several experiments in DE-GAIL, $P\text{(expert)}$ gravitates exceedingly close to one.

Meanwhile, degenerated discriminator behaviors in SD3-GAIL (left of Fig.\,\ref{disc}) are consistent with the cases where returns are zero ($r\rightarrow 0$) in Fig.\,\ref{fig:4algo_3env_11seed}. In comparison, the gradients of PPO-GAIL maintain their training stability.

\begin{figure*}[ht]
\centering
\begin{minipage}{0.32\linewidth}
\includegraphics[width=\textwidth]{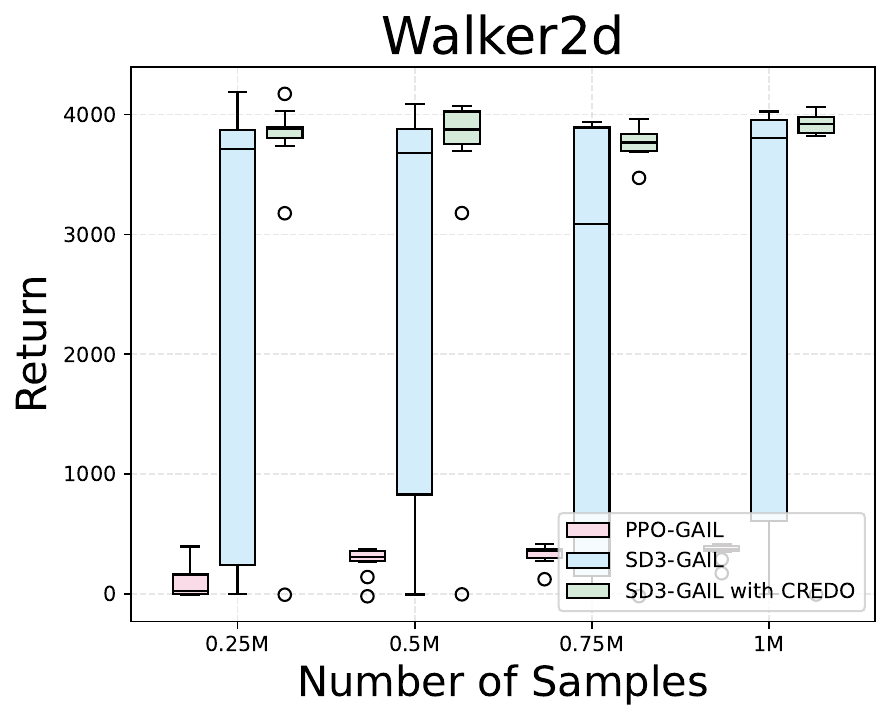}
\end{minipage}
\begin{minipage}{0.32\linewidth}
\includegraphics[width=\textwidth]{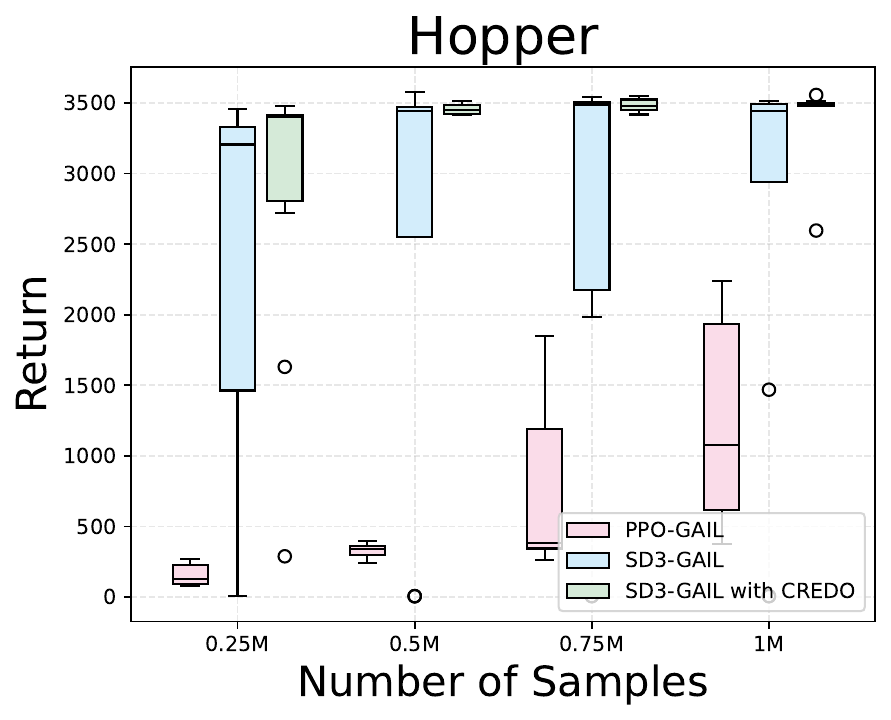}
\end{minipage}
\begin{minipage}{0.32\linewidth}
\includegraphics[width=\textwidth]{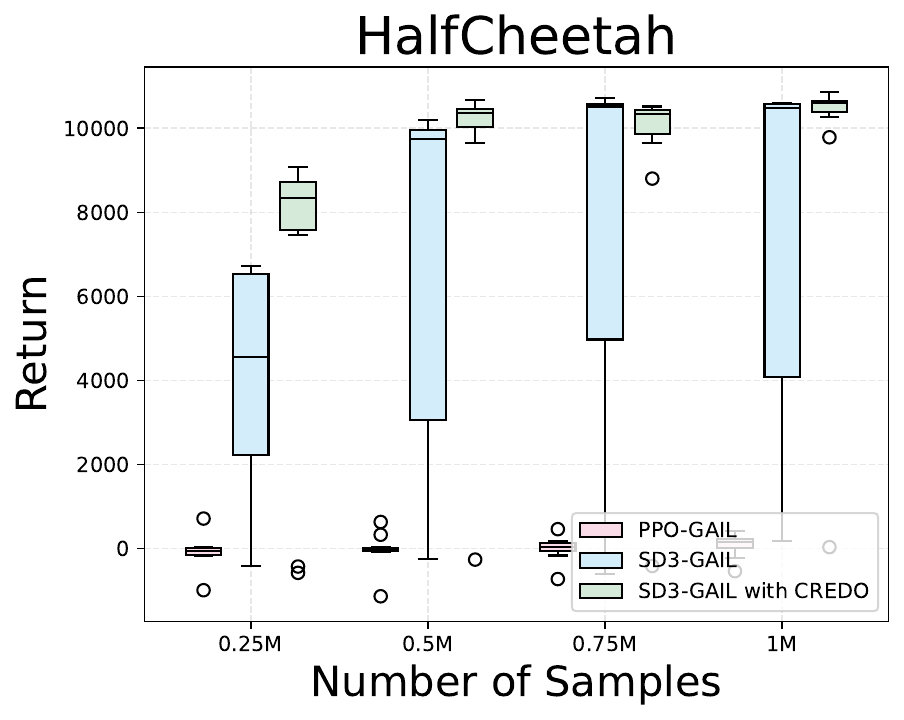}
\end{minipage}
\caption{Box-plot of PPO-GAIL, SD3-GAIL, and SD3-GAIL with CREDO in three environments. The clipping technique significantly enhances the training stability of SD3-GAIL, resulting in a high successful rate in terms of convergence.}
\label{fig:boxplot}
\end{figure*}

\subsection{Relieving Exploding Gradients with Reward 
Modification\label{section_relieving}}

The AIRL~\cite{kostrikov2019discriminator} modified the reward function $r_{2}(s_{t},a_{t})=\log( D(s_{t},a_{t}))-\log(1-D(s_{t},a_{t}))$ into DE-GAIL. This reward function can empirically mitigate the training instability of DE-GAIL. Here we give a theory that supports their experiments and corroborate our analysis. 

The reward function of AIRL is defined as a combination reward function (CR)~\cite{wang2021reward}. For convenience, DE-GAIL with PLR and CR are called PLR-DE-GAIL and CR-DE-GAIL, respectively. We study whether CR-DE-GAIL can have a lower probability of gradient explosion compared to PLR-DE-GAIL. Note that Theorem~\ref{optimal_theorem} shows that the policy disparity causes the gradient explosion. Unlike the discriminator, which can be defined within a finite interval of $(0,1)$, the expert-imitator policy disparity is vaguely defined. Therefore, the following proposition provides a concrete understanding of policy disparity.

\begin{proposition}
When the discriminator achieves its optimum $D^{\ast}(s,a)$ in Eq. (\ref{D_ast}), we have
\begin{align*}
D^{\ast}&(s_{t},a_{t})\approx 1 \Leftrightarrow \\
&h(s_{t})~\text{unequaled}~a_{t}~\text{under the event $\Xi$ in Eq. (\ref{event})}. 
\end{align*}
\label{D_mismatch}
\end{proposition}

\noindent \textbf{Proof~}
See Supplementary Material B.1.\\

Proposition \ref{D_mismatch} indicates that exploding gradients can either depend on the distance between the discriminator's value and value 1, or the degree of $r_i(s_{t},a_{t}), \text{where}~ i=1, 2$ that goes to infinity. This is due to the monotonicity of both $r_{1}(s_{t},a_{t})$ and $r_{2}(s_{t},a_{t})$. As $D(s_{t},a_{t})\approx 1$, we obtain 
\begin{align*}
r_{1}(s_{t},a_{t})\approx \infty ~\text{and}~r_{2}(s_{t},a_{t})\approx \infty.
\end{align*}

Intuitively, we want to prevent exploding gradients. As such, we make the constraints that $r_i(s_{t},a_{t})\leq c,~i=1,2$, for some appropriate constant $c$. In contrast, the outliers of the discriminator can also be characterized as $r_i(s_{t},a_{t})> c$ for $i=1,2$, which represents the situation of gradient explosion. We define such a state as follows:

\begin{definition}
When the discriminator achieves its optimum $D^{\ast}(s,a)$ in Eq. (\ref{D_ast}), the outliers of the discriminator are defined in $[\alpha,1]$ such that $r_{1}(s_{t},a_{t})\geq c$. Similarly, under the same upper bound $c$, the outliers of the discriminator are defined in $[\beta,1]$ for $r_{2}(s_{t},a_{t})$.
\label{alpha}
\end{definition}

We note that the training process will suffer from exploding gradients when the discriminator falls into the $[\alpha,1]$ range. The next proposition describes how to relieve the gradient explosion in CR-DE-GAIL. 

\begin{proposition}
When the discriminator achieves its optimum $D^{\ast}(s,a)$ in Eq. (\ref{D_ast}), we have $\beta \geq \alpha$. 
\label{threshold}
\end{proposition}

\noindent \textbf{Proof~}
See Supplementary Material B.2.\\

Proposition \ref{threshold} reveals that the discriminator in CR-DE-GAIL exhibits a smaller interval of outliers than that in PLR-DE-GAIL, which decreases the probability of gradient explosion. 

\subsection{Clipping Reward of Discriminator Outlier\label{section_clipping}}

The analysis from the preceding section demystified the phenomenon of gradient explosion in GAIL. This examination uncovered that the intrinsic limitations of DE-GAIL can occasionally lead to inevitable divergence. Nonetheless, when DE-GAIL does converge, its data efficiency notably surpasses that of ST-GAIL. As a result, we are driven to develop a robust approach that identifies and alleviates the gradient explosion issue in DE-GAIL while preserving its admirable training efficiency.

Building on the insights from Proposition \ref{D_mismatch}, we highlighted the pivotal role of the reward function in the gradient explosion issue. Inspired by this understanding, we aim to mitigate the likelihood of exploding gradients in DE-GAIL. To achieve this, we apply a clipping mechanism to the rewards that align with the discriminator's outliers in DE-GAIL, enforcing constraints such that $r(s,a)\leq c$ for some appropriate constant $c$ to reduce the outliers. We term this clipping strategy CREDO, an acronym for Clipping REward of Discriminator Outlier. 

It's worth noting that our method is versatile and can be applied across all DE-GAIL algorithms. We selected SD3-GAIL as our baseline and integrated CREDO owing to its standout performance. To maintain consistency and to showcase the resilience of our approach, we performed 11 independent experiments for each methodology, ensuring uniform training settings for all, except for the random seed. We maintained identical hyperparameters across all experiments. In particular, we set the update frequency at 64 and established a threshold $c = 5$. Additional training details, including pseudocode and hyperparameter settings, and additional
experimental results in Humanoid-v2 and Ant-v2 are shown in Supplementary Material F.

The experimental results are illustrated in Figure~\ref{fig:boxplot}. We employ box plots to showcase the returns at various training phases, gauged by the number of samples. Notably, the blue box plot, symbolizing the vanilla SD3-GAIL, displays outlier points with zero return values. These points correspond to experiments that failed to converge, due to gradient explosion. In contrast, the integration of CREDO into SD3-GAIL (as represented by the green box plot) significantly reduces the box plot's span and reduces the outliers. Such outcomes are consistently observed across all environments. When juxtaposed with the ST-GAIL approach, our CREDO method boasts data efficiency that is an order of magnitude higher than that of PPO-GAIL.

\section{Conclusion}
This paper delves into the issue of gradient explosion in Generative Adversarial Imitation Learning (GAIL). The journey begins by examining the existence of divergence in GAIL. Among the two types of GAIL, namely deterministic policy and stochastic policy, we observe that the former has a non-negligible probability of divergence, whereas the latter exhibits successful convergence. To gain an in-depth comprehension of this phenomenon, we analyze it from a theoretical standpoint, explicitly considering the structure of the reward function. Subsequently, we introduce an example featuring a modified reward function, demonstrating that such alterations can effectively mitigate the gradient explosion issue. To further alleviate this problem in DE-GAIL, we propose a novel technique, the efficacy of which is substantiated through experimental evidence. Overall, our analysis of exploding gradients fosters a new understanding of GAIL in terms of training schemes. 

\section*{Acknowledgments}
This work was supported in part by the Key Technologies Research and Development Program under Grant 2021ZD0140300, the National Natural Science Foundation of China under Grants 12301351, 62225308 and 11771276, the Shanghai Sailing Program, Shanghai Association for Science and Technology under Grant 21YF1413500 \enquote{Shuguang Program} under Grant 20SG40 supported by Shanghai Education Development Foundation and Shanghai Municipal Education Commission, and the Science and Technology Innovation Action Plan of Shanghai under Grant 22511105400.

\bibliography{aaai24}
\clearpage

\appendix

\begin{center}
    \Large{\textbf{Supplementary Material for\\ ``Exploring Gradient Explosion in Generative Adversarial Imitation Learning: \\A Probabilistic Perspective''}}
\end{center}

\section{Analysis of Exploding Gradients in DE-GAIL}
\subsection{Proof of Theorem 1}
\textbf{Theorem 1.} \emph{Let $\pi_{h}(\cdot|s)$ be the Gaussian stochastic policy with mean $h(s)$ and covariance $\bm{\Sigma}$. When the discriminator achieves its optimum $D^{\ast}(s,a)$ in Eq. (9), the gradient estimator of the policy loss with respect to the policy’s parameter $h$ satisfies $\|\hat{\nabla}_{h}D_{\rm JS}(\rho^{\pi_{h}},\rho^{\pi_{\rm E}})\|_2\rightarrow \infty$ with a probability of at least $\Pr(\|\bm{\Sigma}^{-1}(a_{t}-h(s_{t}))\|_2\geq C~\text{for any } C>0)$ as $\bm{\Sigma}\rightarrow \bm{0}$, where 
\begin{align*}
\hat{\nabla}_{h}D_{\rm JS}(\rho^{\pi_{h}},\rho^{\pi_{\rm E}})
=&\frac{H_h \Delta\left(T^{\top} \mathbf{d}^{h}\right) (\mathrm{I}-\gamma \mathrm{P} \Pi_h)^{-1}\mathbf{e}_{s_t,a_t}}{2\rho^{\pi_{\rm E}}(s_t,a_t)}\cdot \notag\\
&~~~~~~~~~~~~\log\frac{2\rho^{\pi_{h}}(s_t,a_t)}{\rho^{\pi_{h}}(s_t,a_t)+\rho^{\pi_{\rm E}}(s_t,a_t)},
\end{align*}
and $\left(\mathrm{H}_h\right)_{ s a}
=\pi_{h}(a|s)\kappa(s,\cdot)\bm{\Sigma}^{-1}(a-h(s))$, $\Delta(\cdot)$ maps a vector to a diagonal matrix with its elements on the main diagonal, $\mathbf{e}_{s_t,a_t}=[0,\cdots,\mathop{1}_{s_t,a_t},\cdots,0]^{\top} \in \mathbb{R}^{|\mathcal{S}||\mathcal{A}|\times 1}$. 
}

\begin{proof}
Through importance sampling which transfers the learned state-action distribution to the expert demonstration distribution, the JS divergence can be rewritten from the definition in Eq. (5) as
\begin{align}
&~~~~D_{\rm JS}(\rho^{\pi_{h}},\rho^{\pi_{\rm E}}) \notag\\
&=\frac{1}{2}D_{\rm KL}(\rho^{\pi_{h}},\frac{\rho^{\pi_{h}}+\rho^{\pi_{\rm E}}}{2}) + \frac{1}{2}D_{\rm KL}(\rho^{\pi_{\rm E}},\frac{\rho^{\pi_{h}}+\rho^{\pi_{\rm E}}}{2}) \notag\\
&=\frac{1}{2}\mathbb{E}_{(s,a)\sim \mathcal{D}}\left[ \log\frac{2\rho^{\pi_h}(s,a)}{\rho^{\pi_h}(s,a)+\rho^{\pi_{\rm E}}(s,a)} \right]\notag\\
&~~~~~+ \frac{1}{2}\mathbb{E}_{(s,a)\sim \mathcal{D}^{\star}}\left[ \log\frac{2\rho^{\pi_{\rm E}}(s,a)}{\rho^{\pi_h}(s,a)+\rho^{\pi_{\rm E}}(s,a)} \right] \notag\\
&=\frac{1}{2}\mathbb{E}_{(s,a)\sim \mathcal{D}^{\star}}\bigg[ \frac{\rho^{\pi_h}(s,a)}{\rho^{\pi_{\rm E}}(s,a)}\log\frac{2\rho^{\pi_h}(s,a)}{\rho^{\pi_h}(s,a)+\rho^{\pi_{\rm E}}(s,a)}\notag\\
&~~~~~+\log\frac{2\rho^{\pi_{\rm E}}(s,a)}{\rho^{\pi_h}(s,a)+\rho^{\pi_{\rm E}}(s,a)} \bigg], 
\tag{16}
\label{JS_to_expert}
\end{align}
where $\mathcal{D}^{\star}$ and $\mathcal{D}$ denote the expert demonstration and the replay buffer of $\pi_{h}$ respectively. Then we can approximate the gradient of Eq. (\ref{JS_to_expert}) with respect to $h$ with
\begin{align}
&~~~~\hat{\nabla}_{h}D_{\rm JS}(\rho^{\pi_{h}},\rho^{\pi_{\rm E}}) \notag\\
&=\frac{1}{2}\nabla_{h}\bigg( \frac{\rho^{\pi_h}(s_{t},a_{t})}{\rho^{\pi_{\rm E}}(s_{t},a_{t})}\log\frac{2\rho^{\pi_h}(s_{t},a_{t})}{\rho^{\pi_h}(s_{t},a_{t})+\rho^{\pi_{\rm E}}(s_{t},a_{t})}\notag\\
&~~~~+\log \frac{2\rho^{\pi_{\rm E}}(s_{t},a_{t})}{\rho^{\pi_h}(s_{t},a_{t})+\rho^{\pi_{\rm E}}(s_{t},a_{t})} \bigg) \notag\\
&=\frac{1}{2}\bigg( \frac{\nabla_{h}\rho^{\pi_{h}}(s_t,a_t)}{\rho^{\pi_{\rm E}}(s_t,a_t)}\log\frac{2\rho^{\pi_{h}}(s_t,a_t)}{\rho^{\pi_{h}}(s_t,a_t)+\rho^{\pi_{\rm E}}(s_t,a_t)} \notag\\
&~~~~+\frac{\rho^{\pi_{h}}(s_{t},a_{t})}{\rho^{\pi_{\rm E}}(s_{t},a_{t})}\cdot \frac{\rho^{\pi_h}(s_{t},a_{t})+\rho^{\pi_{\rm E}}(s_{t},a_{t})}{2\rho^{\pi_{h}}(s_{t},a_{t})}\cdot \notag\\
&~~~~\big(\frac{2\nabla_{h}\rho^{\pi_{h}}(s_t,a_t)\big( \rho^{\pi_h}(s_{t},a_{t})+\rho^{\pi_{\rm E}}(s_{t},a_{t}) \big)}{\big( \rho^{\pi_h}(s_{t},a_{t})+\rho^{\pi_{\rm E}}(s_{t},a_{t}) \big)^{2}}  \notag\\
&~~~~~~~~~~~~~~~~~~~~~~~~~~~~~~
-\frac{2\rho^{\pi_{h}}(s_{t},a_{t})\nabla_{h}\rho^{\pi_{h}}(s_t,a_t)}{\big(\rho^{\pi_h}(s_{t},a_{t})+\rho^{\pi_{\rm E}}(s_{t},a_{t}) \big)^{2}} \big) \notag\\
&~~~~-\frac{\rho^{\pi_h}(s_{t},a_{t})+\rho^{\pi_{\rm E}}(s_{t},a_{t})}{2\rho^{\pi_{\rm E}}(s_{t},a_{t})}\cdot \notag\\
&~~~~~~~~~~~~~~~~~~~~~~~~~~~~~~\frac{2\rho^{\pi_{\rm E}}(s_{t},a_{t})\nabla_{h}\rho^{\pi_{h}}(s_t,a_t)}{\big( \rho^{\pi_h}(s_{t},a_{t})+\rho^{\pi_{\rm E}}(s_{t},a_{t}) \big)^{2}} \bigg) \notag\\
&=\frac{\nabla_{h}\rho^{\pi_{h}}(s_t,a_t)}{2\rho^{\pi_{\rm E}}(s_t,a_t)}\log\frac{2\rho^{\pi_{h}}(s_t,a_t)}{\rho^{\pi_{h}}(s_t,a_t)+\rho^{\pi_{\rm E}}(s_t,a_t)}
\tag{17}
\label{JS_gradient_estimator}
\end{align}

Let the derivative matrix $\Upsilon$ of the state-action distribution w.r.t. the
policy parameters $h$ to be $\Upsilon_{s a}=\nabla_{h}\rho^{\pi_h}(s, a)$. By the fact that 
\begin{align*}
\Upsilon=H_h \Delta\left(T^{\top} \mathbf{d}^{h}\right) \Psi_h^{-1}, 
\end{align*}
where $\Psi_h =\mathrm{I}-\gamma \mathrm{P} \Pi_h$ and  
$\Delta(\cdot)$ maps a vector to a diagonal matrix with its elements on the main diagonal \cite{wen2021characterizing}, we have
\begin{align}
\nabla_{h}\rho^{\pi_{h}}(s,a)=\Upsilon_{s a}=
H_h \Delta\left(T^{\top} \mathbf{d}^{h}\right) \Psi_h^{-1}\mathbf{e}_{s,a}, 
\tag{18}
\label{gradient_rho}
\end{align}
where $\mathbf{e}_{s,a}=[0,\cdots,\mathop{1}_{s,a},\cdots,0]^{\top} \in \mathbb{R}^{|\mathcal{S}||\mathcal{A}|\times 1}$. 
Eq. (\ref{JS_gradient_estimator}) can be shown that 
\begin{align*}
&~~~~~\|\hat{\nabla}_{h}D_{\rm JS}(\rho^{\pi_{h}},\rho^{\pi_{\rm E}})\|_2 \notag\\
&=\big\|\frac{H_h \Delta\left(T^{\top} \mathbf{d}^{h}\right) (\mathrm{I}-\gamma \mathrm{P} \Pi_h)^{-1}\mathbf{e}_{s_t,a_t}}{2\rho^{\pi_{\rm E}}(s_t,a_t)}\cdot \notag \\
&~~~~~~~~~~~~~~~~~~~~~~~~~~~~~~~~~~~~~~~~\log\frac{2\rho^{\pi_{h}}(s_t,a_t)}{\rho^{\pi_{h}}(s_t,a_t)+\rho^{\pi_{\rm E}}(s_t,a_t)} \big\|_{2}.
\end{align*}
Note that
\begin{align}
\left(\mathrm{H}_h\right)_{ s a}&=\nabla_{h}\pi_{h}(a|s)=\pi_{h}(a|s)\nabla_{h}\log \pi_{h}(a|s)\notag\\
&=\pi_{h}(a|s)\kappa(s,\cdot)\bm{\Sigma}^{-1}(a-h(s)), 
\tag{19}
\label{gradient_pi}
\end{align}
then it follows that $\|\hat{\nabla}_{h}D_{\rm JS}(\rho^{\pi_{h}},\rho^{\pi_{\rm E}})\|_2\rightarrow \infty $ with a probability of at least $\Pr(\|\bm{\Sigma}^{-1}(a_{t}-h(s_{t}))\|_2\geq C~\text{for any } C>0)$ as $\bm{\Sigma}\rightarrow \bm{0}$. 
\end{proof}

\subsection{Proof of Corollary 1}
\textbf{Corollary 1.} \emph{Let $\pi_{h}(\cdot|s)$ be the Gaussian stochastic policy with mean $h(s)$ and covariance $\bm{\Sigma}$. When the discriminator is in the format of Eq. (13), i.e., $\Tilde{D}(s,a)\in (0,1)$, the gradient estimator of the policy loss with respect to the policy’s parameter $h$ satisfies 
\begin{align*}
\left\| \hat{\nabla}_{h}\left( \mathbb{E}_{\mathcal{D}^{\star}}[\log \tilde{D}(s,a) ]+\mathbb{E}_{\mathcal{D}}[\log(1-\tilde{D}(s,a))] \right) \right\|_{2}\rightarrow \infty
\end{align*}
with a probability of $\Pr(\|\bm{\Sigma}^{-1}(a_{t}-h(s_{t}))\|_2\geq C~\text{for any } C>0)$ as $\bm{\Sigma}\rightarrow \bm{0}$, where $\mathcal{D}^{\star}$ and $\mathcal{D}$ denote the expert demonstration and the replay buffer of $\pi_{h}$ respectively, 
\begin{align*}
&~~~~\hat{\nabla}_{h}\left( \mathbb{E}_{\mathcal{D}^{\star}}[\log(\tilde{D}(s,a))]+\mathbb{E}_{\mathcal{D}}[\log(1-\tilde{D}(s,a))] \right) \\
&=\frac{H_h \Delta\left(T^{\top} \mathbf{d}^{h}\right) (\mathrm{I}-\gamma \mathrm{P} \Pi_h)^{-1}\mathbf{e}_{s_t,a_t}}{\rho^{\pi_{\rm E}}(s_{t},a_{t})}\cdot\\
&~~~~~~~~~~~~\log \frac{(1-\epsilon_{2})\rho^{\pi_{h}}(s_t,a_t)}{(1+\epsilon_{1})\rho^{\pi_{\rm E}}(s_{t},a_{t})+(1-\epsilon_{2})\rho^{\pi_{h}}(s_t,a_t)} \\
&~~~~+\frac{(\epsilon_{1}+\epsilon_{2})\nabla_{h}\rho^{\pi_{h}}(s_t,a_t)}{(1+\epsilon_{1})\rho^{\pi_{\rm E}}(s_{t},a_{t})+(1-\epsilon_{2})\rho^{\pi_{h}}(s_{t},a_{t})},
\end{align*}
and $\left(\mathrm{H}_h\right)_{ s a}
=\pi_{h}(a|s)\kappa(s,\cdot)\bm{\Sigma}^{-1}(a-h(s))$, $\Delta(\cdot)$ maps a vector to a diagonal matrix with its elements on the main diagonal, $\mathbf{e}_{s_t,a_t}=[0,\cdots,\mathop{1}_{s_t,a_t},\cdots,0]^{\top} \in \mathbb{R}^{|\mathcal{S}||\mathcal{A}|\times 1}$. 
}

\begin{proof}
Referring to the proof strategy of Theorem 1, the learned state-action distribution can be transferred to the expert demonstration distribution by importance sampling. Thus when the discriminator achieves its regularity $\Tilde{D}(s,a)$, we can write the policy objective from the optimization problem in Eq. (3) as 
\begin{align}
&~~~~\mathbb{E}_{(s,a)\sim \mathcal{D}^{\star}}[\log(\tilde{D}(s,a))]+\mathbb{E}_{(s,a)\sim \mathcal{D}}[\log(1-\tilde{D}(s,a))] \notag\\
&=\mathbb{E}_{(s,a)\sim \mathcal{D}^{\star}}\left[\log \frac{(1+\epsilon_{1})\rho^{\pi_{\rm E}}(s_{t},a_{t})}{(1+\epsilon_{1})\rho^{\pi_{\rm E}}(s_{t},a_{t})+(1-\epsilon_{2})\rho^{\pi}(s_{t},a_{t})} \right] \notag\\
&~~+\mathbb{E}_{(s,a)\sim \mathcal{D}}\left[ \log \frac{(1-\epsilon_{2})\rho^{\pi}(s_{t},a_{t})}{(1+\epsilon_{1})\rho^{\pi_{\rm E}}(s_{t},a_{t})+(1-\epsilon_{2})\rho^{\pi}(s_{t},a_{t})} \right] \notag\\
&=\mathbb{E}_{(s,a)\sim \mathcal{D}^{\star}}\bigg[ \log \frac{(1+\epsilon_{1})\rho^{\pi_{\rm E}}(s,a)}{(1+\epsilon_{1})\rho^{\pi_{\rm E}}(s,a)+(1-\epsilon_{2})\rho^{\pi_{h}}(s,a)} \notag\\
&~~+\frac{\rho^{\pi_{h}}(s,a)}{\rho^{\pi_{\rm E}}(s,a)}\log \frac{(1-\epsilon_{2})\rho^{\pi_{h}}(s,a)}{(1+\epsilon_{1})\rho^{\pi_{\rm E}}(s,a)+(1-\epsilon_{2})\rho^{\pi_{h}}(s,a)} \bigg]. 
\tag{20}
\label{any_D_policy_objective}
\end{align}
Then the gradient of Eq. (\ref{any_D_policy_objective}) can be approximated with
\begin{align}
&~~~~\hat{\nabla}_{h}\left( \mathbb{E}_{(s,a)\sim \mathcal{D}^{\star}}[\log(\tilde{D}(s,a))]+\mathbb{E}_{\mathcal{D}}[\log(1-\tilde{D}(s,a))] \right) \notag\\
&=\nabla_{h}\bigg( \log \frac{(1+\epsilon_{1})\rho^{\pi_{\rm E}}(s_{t},a_{t})}{(1+\epsilon_{1})\rho^{\pi_{\rm E}}(s_{t},a_{t})+(1-\epsilon_{2})\rho^{\pi_{h}}(s_{t},a_{t})} \notag\\
&~~~~+\frac{\rho^{\pi_{h}}(s_{t},a_{t})}{\rho^{\pi_{\rm E}}(s_{t},a_{t})}\cdot \notag\\
&~~~~~~~~~~~~~~~~~\log \frac{(1-\epsilon_{2})\rho^{\pi_{h}}(s_{t},a_{t})}{(1+\epsilon_{1})\rho^{\pi_{\rm E}}(s_{t},a_{t})+(1-\epsilon_{2})\rho^{\pi_{h}}(s_{t},a_{t})} \bigg) \notag\\
&=-\frac{(1+\epsilon_{1})\rho^{\pi_{\rm E}}(s_{t},a_{t})+(1-\epsilon_{2})\rho^{\pi_{h}}(s_{t},a_{t})}{(1+\epsilon_{1})\rho^{\pi_{\rm E}}(s_{t},a_{t})}\cdot \notag\\
&~~~~\frac{(1+\epsilon_{1})(1-\epsilon_{2})\rho^{\pi_{\rm E}}(s_{t},a_{t})\nabla_{h}\rho^{\pi_{h}}(s_t,a_t)}{\big( (1+\epsilon_{1})\rho^{\pi_{\rm E}}(s_{t},a_{t})+(1-\epsilon_{2})\rho^{\pi_{h}}(s_{t},a_{t}) \big)^{2}} \notag\\
&~~~~+\frac{\nabla_{h}\rho^{\pi_{h}}(s_t,a_t)}{\rho^{\pi_{\rm E}}(s_{t},a_{t})}\cdot \notag\\
&~~~~~~~~~~~~~~~~~~~~~\log \frac{(1-\epsilon_{2})\rho^{\pi_{h}}(s_t,a_t)}{(1+\epsilon_{1})\rho^{\pi_{\rm E}}(s_{t},a_{t})+(1-\epsilon_{2})\rho^{\pi_{h}}(s_t,a_t)} \notag\\
&~~~~+\frac{\rho^{\pi_{h}}(s_{t},a_{t})}{\rho^{\pi_{\rm E}}(s_{t},a_{t})}\cdot \notag\\
&~~~~~~~~~~~~~~~~~~~~~\frac{(1+\epsilon_{1})\rho^{\pi_{\rm E}}(s_{t},a_{t})+(1-\epsilon_{2})\rho^{\pi_{h}}(s_{t},a_{t})}{(1-\epsilon_{2})\rho^{\pi_{h}}(s_{t},a_{t})} \notag\\
&~~~~\cdot \frac{(1-\epsilon_{2})(1+\epsilon_{1})\rho^{\pi_{\rm E}}(s_{t},a_{t})\nabla_{h}\rho^{\pi_{h}}(s_t,a_t)}{\big( (1+\epsilon_{1})\rho^{\pi_{\rm E}}(s_{t},a_{t})+(1-\epsilon_{2})\rho^{\pi_{h}}(s_{t},a_{t}) \big)^{2}} \notag\\
&=\frac{\nabla_{h}\rho^{\pi_{h}}(s_t,a_t)}{\rho^{\pi_{\rm E}}(s_{t},a_{t})}\cdot \notag\\
&~~~~~~~~~~~~~~~~~~~\log \frac{(1-\epsilon_{2})\rho^{\pi_{h}}(s_{t},a_{t})}{(1+\epsilon_{1})\rho^{\pi_{\rm E}}(s_{t},a_{t})+(1-\epsilon_{2})\rho^{\pi_{h}}(s_{t},a_{t})} \notag\\
&~~~~-\frac{(1-\epsilon_{2})\nabla_{h}\rho^{\pi_{h}}(s_t,a_t)}{(1+\epsilon_{1})\rho^{\pi_{\rm E}}(s_{t},a_{t})+(1-\epsilon_{2})\rho^{\pi_{h}}(s_{t},a_{t})}+\notag\\
&~~~~\frac{(1+\epsilon_{1})\nabla_{h}\rho^{\pi_{h}}(s_t,a_t)}{(1+\epsilon_{1})\rho^{\pi_{\rm E}}(s_{t},a_{t})+(1-\epsilon_{2})\rho^{\pi_{h}}(s_{t},a_{t})} \notag\\
&=\frac{\nabla_{h}\rho^{\pi_{h}}(s_t,a_t)}{\rho^{\pi_{\rm E}}(s_{t},a_{t})}\cdot \notag\\
&~~~~~~~~~~~~~~~~~~~\log\frac{(1-\epsilon_{2})\rho^{\pi_{h}}(s_{t},a_{t})}{(1+\epsilon_{1})\rho^{\pi_{\rm E}}(s_{t},a_{t})+(1-\epsilon_{2})\rho^{\pi_{h}}(s_{t},a_{t})} \notag\\
&~~~~+\frac{(\epsilon_{1}+\epsilon_{2})\nabla_{h}\rho^{\pi_{h}}(s_t,a_t)}{(1+\epsilon_{1})\rho^{\pi_{\rm E}}(s_{t},a_{t})+(1-\epsilon_{2})\rho^{\pi_{h}}(s_{t},a_{t})}. 
\tag{21}
\label{policy_gradient_gail_objective}
\end{align}
Plugging Eq. (\ref{gradient_rho}) and Eq. (\ref{gradient_pi}) into Eq. (\ref{policy_gradient_gail_objective}), when $\|\bm{\Sigma}^{-1}(a_{t}-h(s_{t}))\|_2\geq C~\text{for any } C>0$, we have
\begin{align*}
&\left\| \hat{\nabla}_{h}\left( \mathbb{E}_{(s,a)\sim \mathcal{D}^{\star}}[\log(\tilde{D}(s,a))]+\mathbb{E}_{\mathcal{D}}[\log(1-\tilde{D}(s,a))] \right) \right\|_{2}\notag\\
&~~~~~~~~~~~~~~~~~~~~~~~~~~~~~~~~~~~~~~~~~~~~~~~~~~~~~~~~~~~~~~~~~~~~~~~~~~~~~~~~~~\rightarrow \infty. 
\end{align*}
\end{proof}

\renewcommand{\thefigure}{6}
\begin{figure*}[ht]
\centerline{
\begin{minipage}{0.32\linewidth}
\includegraphics[width=\textwidth]{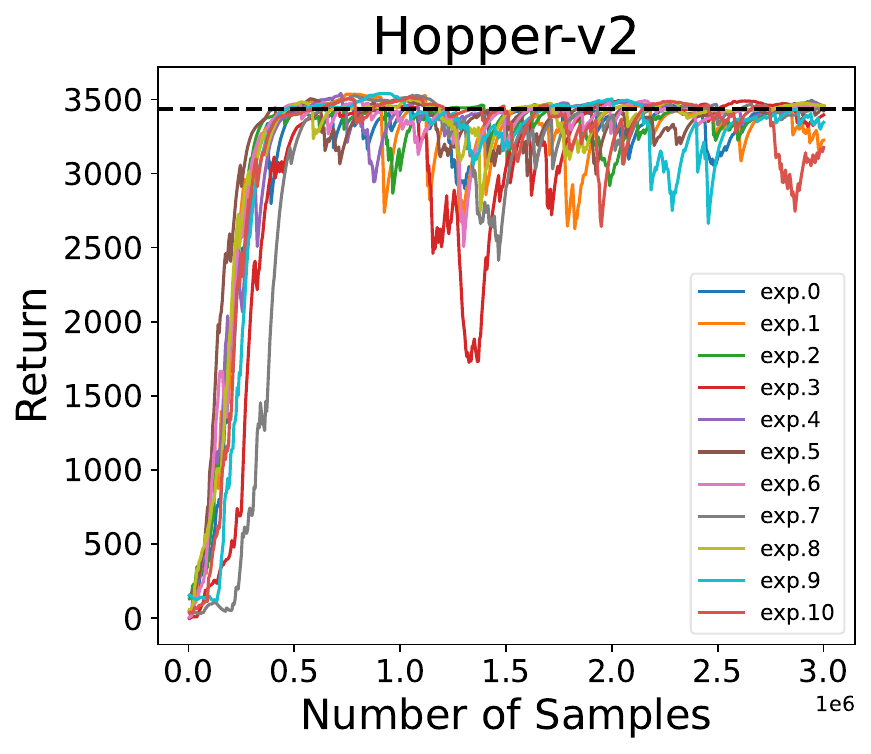}
\end{minipage}
\begin{minipage}{0.32\linewidth}
\includegraphics[width=\textwidth]{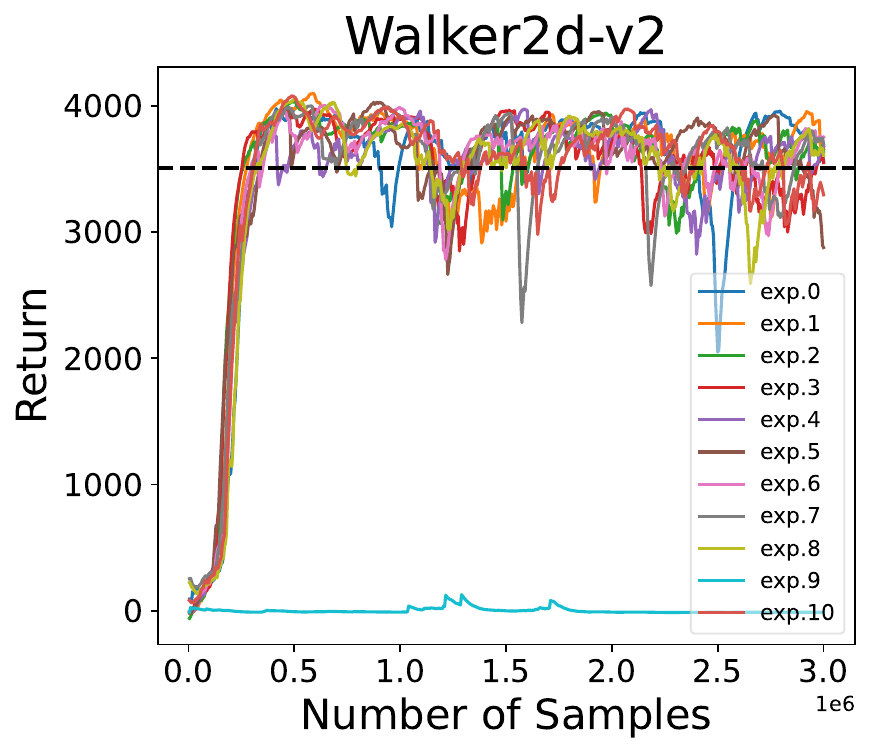}
\end{minipage}
\begin{minipage}{0.32\linewidth}
\includegraphics[width=\textwidth]{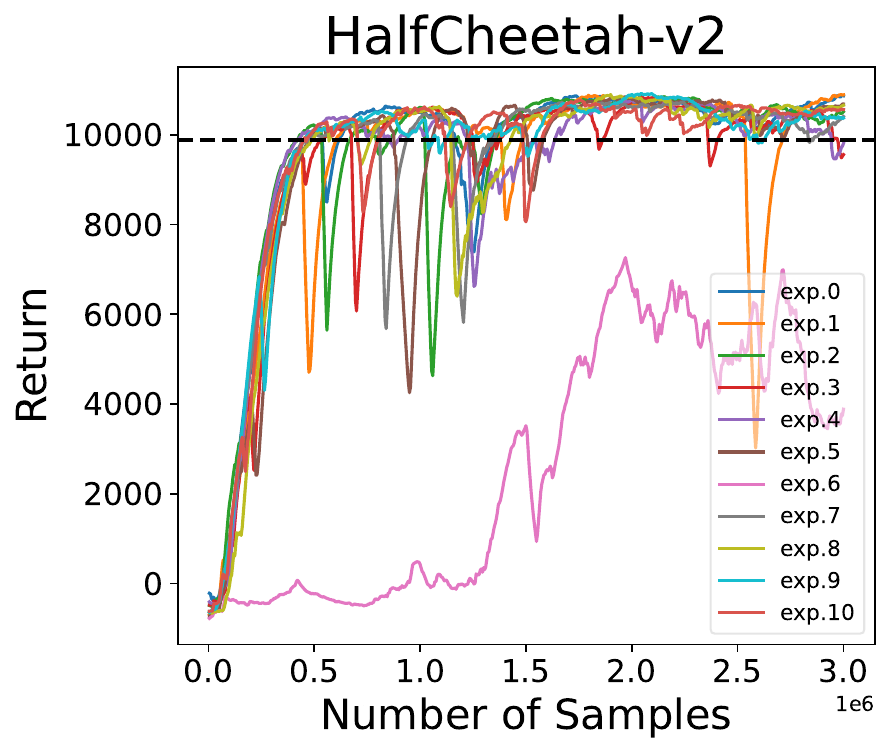}
\end{minipage}}
\caption{SD3-GAIL with CREDO in
three different environments. } 
\label{fig:SD3-GAIL-CREDO}
\end{figure*}

\renewcommand{\thefigure}{7}
\begin{figure*}[ht]
\centerline{
\begin{minipage}{0.32\linewidth}
\includegraphics[width=\textwidth]{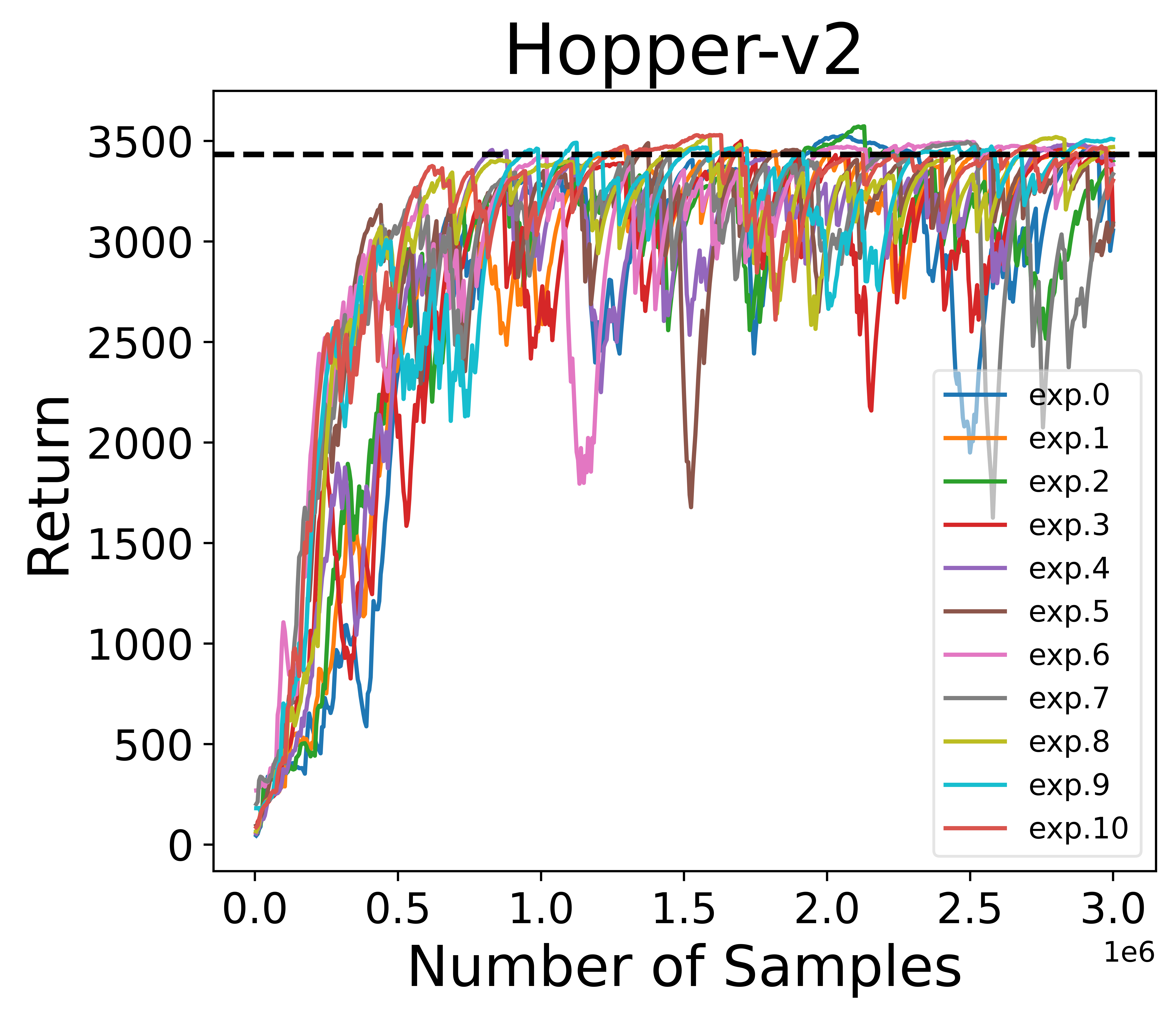}
\end{minipage}
\begin{minipage}{0.32\linewidth}
\includegraphics[width=\textwidth]{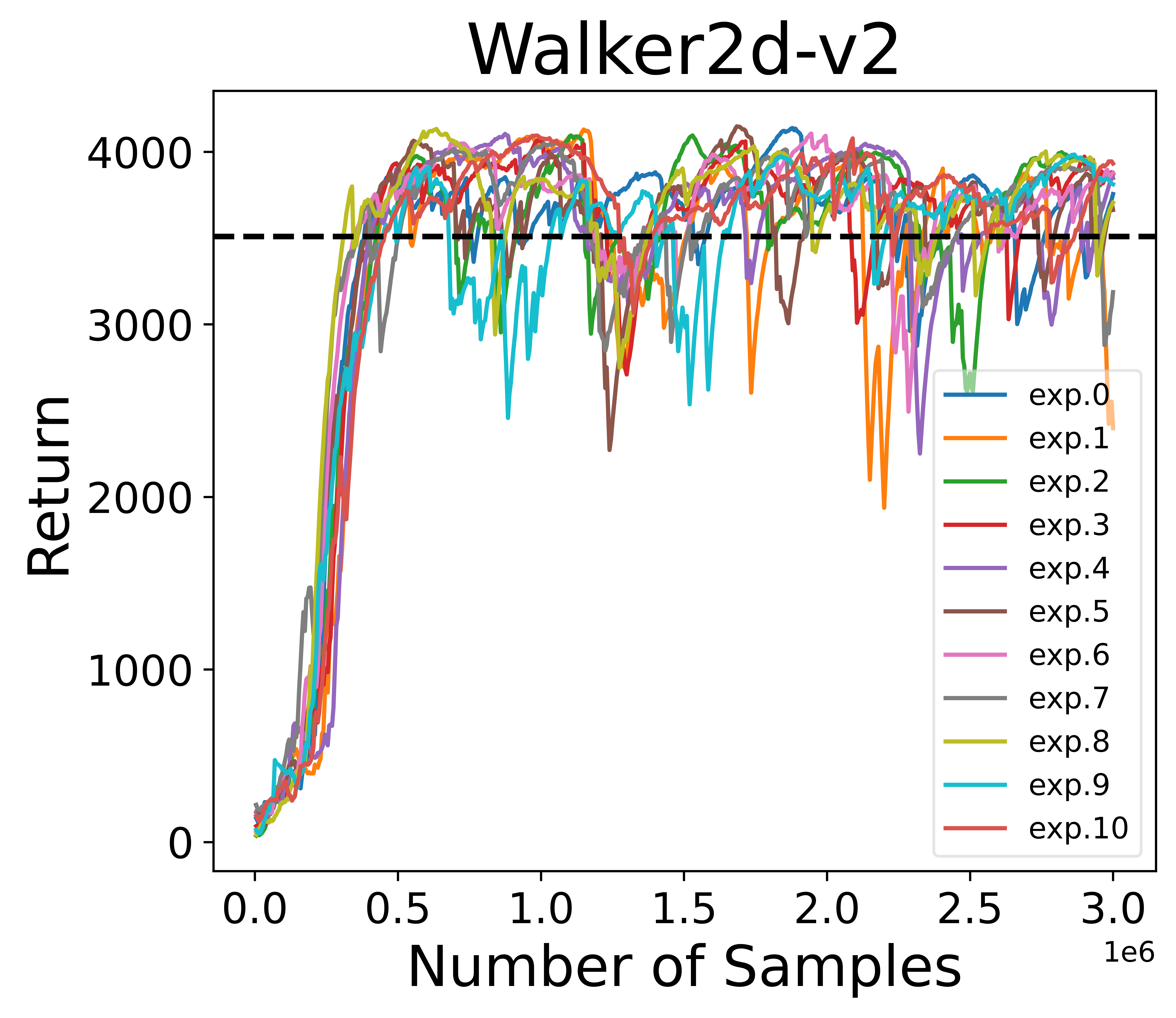}
\end{minipage}
\begin{minipage}{0.32\linewidth}
\includegraphics[width=\textwidth]{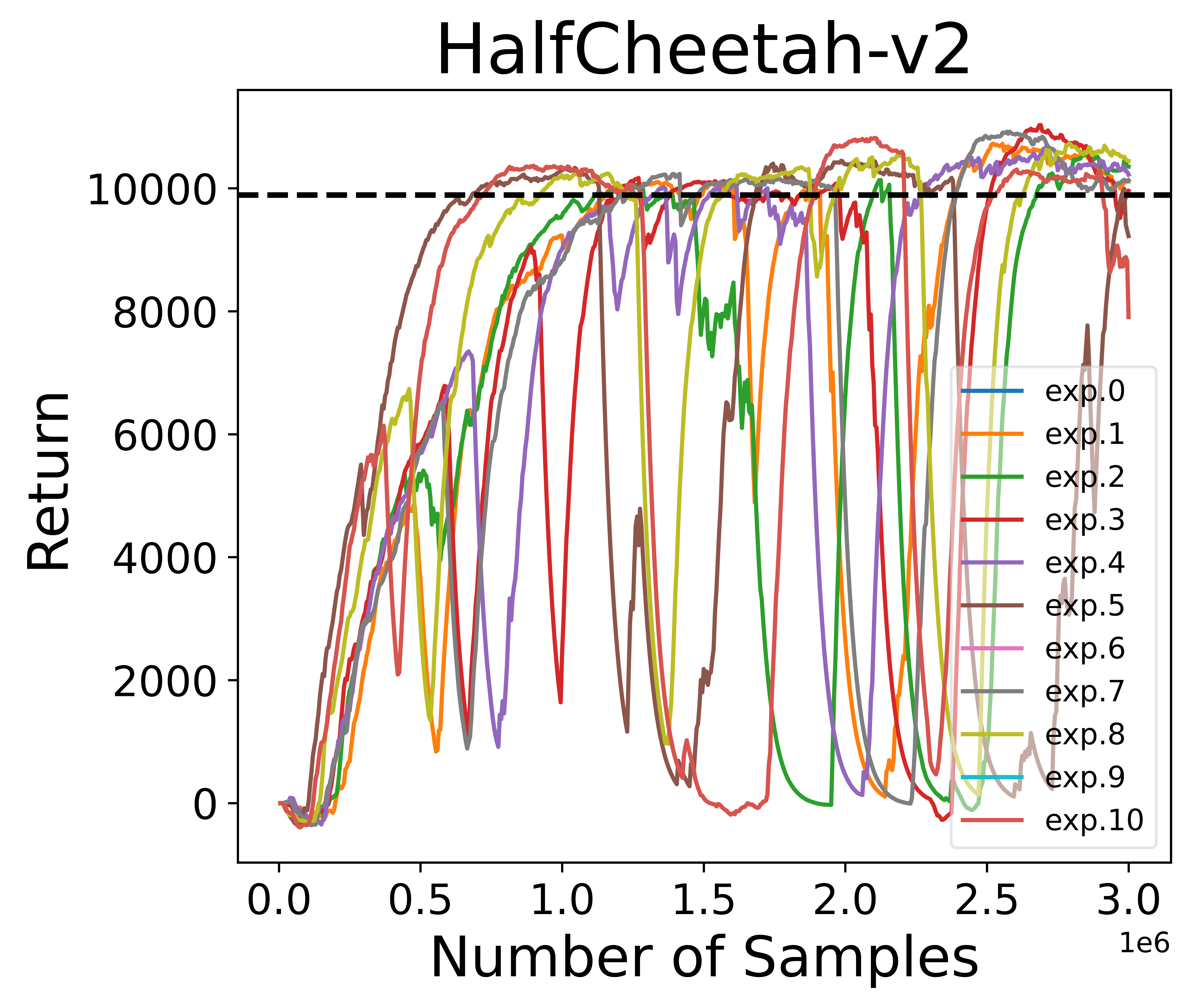}
\end{minipage}}
\caption{TSSG in three different environments. }
\label{fig:TSSG_3env}
\end{figure*}

\section{Analysis of Relieving Exploding Gradients}
\subsection{Proof of Proposition 1}
\textbf{Proposition 1.} \emph{When the discriminator achieves its optimum $D^{\ast}(s,a)$ in Eq. (9), we have
\begin{align*}
D^{\ast}(s_{t},a_{t})\approx 1& \Leftrightarrow h(s_{t})~\text{unequaled}~a_{t}\notag\\
&~~~~~~\text{under the event $\Xi$ in Eq. (11)}.
\end{align*}
}

\begin{proof}
The optimal discriminator of $(s_{t},a_{t})$ can be denoted by
\begin{align*}
D^{\ast}(s_{t},a_{t})=\frac{\rho^{\pi_{\rm E}}(s_{t},a_{t})}{\rho^{\pi_{\rm E}}(s_{t},a_{t})+\rho^{\pi_{h}}(s_{t},a_{t})}.
\end{align*}
We can derive that the necessary and sufficient condition of $D^{\ast}(s_{t},a_{t})\approx 1$ is that $\rho^{\pi_{h}}(s_{t},a_{t})\approx 0$, i.e., $h(s_{t})$ unequaled $a_{t}$ under the event $\Xi$.
\end{proof}

\subsection{Proof of Proposition 2}
\textbf{Proposition 2.} \emph{When the discriminator achieves its optimum $D^{\ast}(s,a)$ in Eq. (9), we have $\beta \geq \alpha$. 
}

\begin{proof}
When $r_{i}(s_t,a_t)=c,~i=1,2$, we obtain $\log \beta-\log(1-\beta)=-\log(1-\alpha)$, which is followed by
\begin{align*}
\beta-\alpha=\frac{\alpha^2-2\alpha+1}{2-\alpha}\geq 0.
\end{align*}
\end{proof} 

\section{Representative Reward Functions}
\begin{table}[H]
\centering
\begin{tabular}{l l}
\hline
\textbf{Reward} & \textbf{Function Shape}\\
\hline
$r_1(s,a)$ & $-\log(1-D(s,a))$ \\ 
$r_2(s,a)$ & $\log D(s,a)-\log(1-D(s,a))$ \\ 
$r_3(s,a)$ & $\log D(s,a)$ \\ 
$r_4(s,a)$ & $D(s,a)$ \\
$r_5(s,a)$ & $e^{D(s,a)}$ \\
$r_6(s,a)$ & $-1/D(s,a)$ \\
$r_7(s,a)$ & $D(s,a)^2$ \\
$r_8(s,a)$ & $\sqrt{D(s,a)}$ \\
\hline
\end{tabular}
\caption{Representative Reward Functions. }
\label{table_2}
\end{table}

\section{Related works}
In imitation learning, where the reward function is inaccessible, GAIL \cite{ho2016generative} is proposed to learn the policy and reward function jointly. It gains significant performance in imitating complex expert policies~\cite{xu2020error,chen2021generative,zhang2022improve,shani2022online}. GAIL frequently employs stochastic policy algorithms, such as Trust Region Policy Optimization \cite{schulman2015trust} (TRPO)-GAIL \cite{ho2016generative}, PPO-GAIL, natural policy gradient (NPG)-GAIL \cite{guan2021when} and TSSG. However, these approaches demand a substantial number of environment samples and extensive training time to develop intricate models for imitation tasks.

To accelerate the GAIL learning process, a natural idea is to use deterministic policy gradients. Sample-efficient adversarial mimic (SAM)~\cite{blonde2019sample} method integrates DDPG into GAIL and adds a penalty on the gradient of the discriminator meanwhile. DGAIL~\cite{zuo2020deterministic} proposed deterministic generative adversarial imitation learning, which combines the modified DDPG with learning from demonstrations to train the generator under the guidance of the discriminator. The reward function in DGAIL is set as $D(s,a)$. TD3 and off-policy training of the discriminator are performed in DAC~\cite{kostrikov2019discriminator} to reduce policy-environment interaction sample complexity by an average factor of 10. The revised reward function in DAC is $\log(D(s,a))-\log(1-D(s,a))$. Notably, these works achieve gratifying results benefiting from modifications to reward functions. The DAC discussed the reward bias in GAIL through a toy example. Different from their work, we provide a comprehensive study on both empirical and theoretical perspectives in terms of gradient explosion in GAIL. 
\renewcommand{\thefigure}{8}
\begin{figure}[t]
\centerline{
\includegraphics[width=0.8\linewidth]{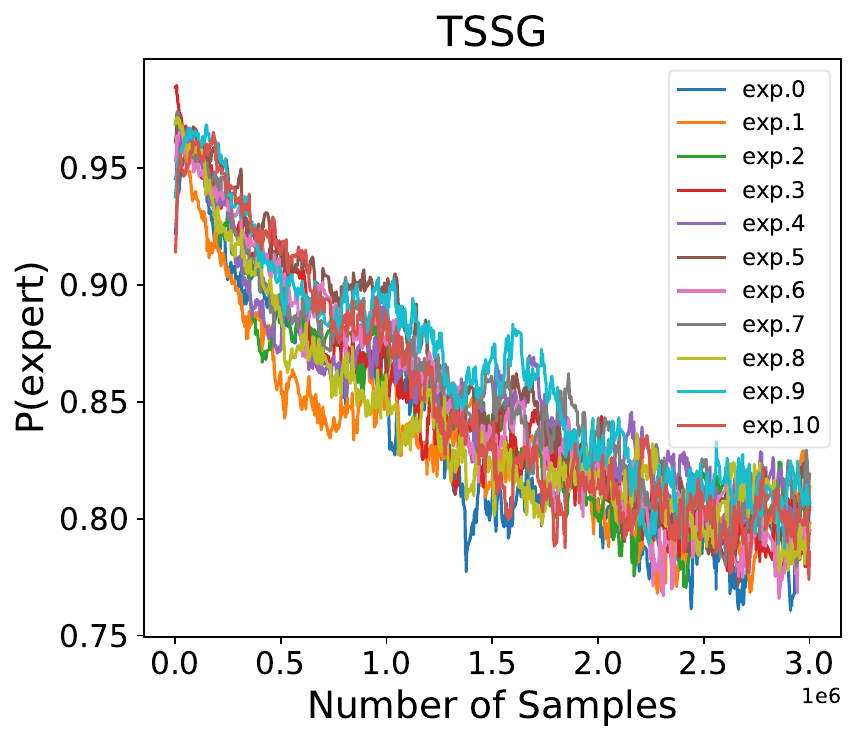}}
\caption{The discriminator of TSSG in Walker2d-v2. }
\label{fig:TSSG_discriminator}
\end{figure}

\renewcommand{\thefigure}{9}
\begin{figure}[t]
\centerline{
\includegraphics[width=0.8\linewidth]{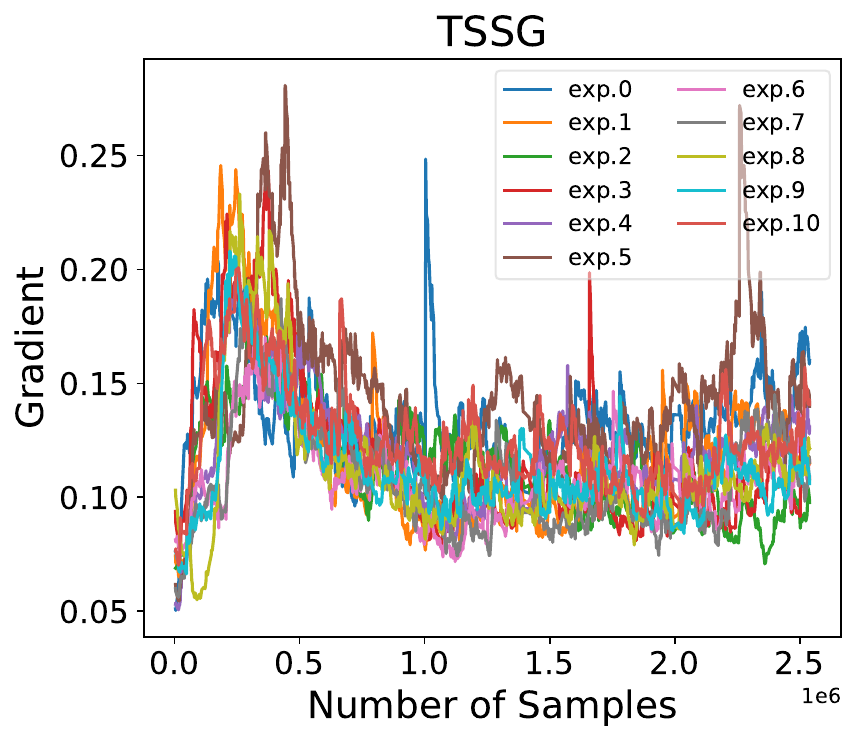}}
\caption{The absolute gradients of TSSG policy networks in Walker2d-v2. }
\label{fig:tssg_gradients}
\end{figure}

\section{Clipping Reward of Discriminator Outlier}
\subsection{The CREDO strategy}
The full CREDO strategy is shown in Algorithm \ref{CREDO}.

\subsection{Proof of CREDO's mitigation of exploding gradients in DE-GAIL}
\textbf{Property.} \emph{CREDO mitigates the exploding gradients in DE-GAIL.}
\begin{proof}
Applying the clipping technique on PLR, $(s_t,a_t)$ in the expert demonstrations $\mathcal{D}^{\star}$ is filtered such that $r(s_t,a_t)<c$, we have
\begin{align*}
D(s_t,a_t) < 1 - e^{-c}. 
\end{align*}
By the definition of the optimal discriminator in Eq. (9), we known that
\begin{align*}
\rho^{\pi_{h}}(s_t,a_t) > \frac{e^{-c}}{1-e^{-c}}\rho^{\pi_{\rm E}}(s_t,a_t), 
\end{align*}
which implies that the imitated action  $h(s_t)=a_t$ with a certain probability. Therefore, the probability of policy disparity $\Pr(\Xi)=\Pr(\|a_{t}-h(s_{t})\|_2\geq C\|\bm{\Sigma}\|_2~\text{for any } C>0)$ is decreased. From Theorem 1, the gradient explosion in DE-GAIL is mitigated with such integration of the reward clipping technique.
\end{proof}

\begin{algorithm}[ht]
\caption{CREDO}
\label{CREDO} 
\begin{algorithmic}[1] 
\STATE \textbf{Input:} Expert demonstrations $\mathcal{D}^{\star}$, a specific reward function $r(s,a)$, two empty datasets $\mathcal{D}$ and $\mathcal{D}_{\rm clip}$, clipping reward threshold $c$.
\STATE Initialize the learned policy $\pi$ and the discriminator $D$. 
\FOR{iteration $0,1,2,\cdots$}
\STATE $\mathcal{D}_{\rm clip}\leftarrow \emptyset$. 
\STATE $\mathcal{D}\leftarrow$ Collect samples using policy $\pi$. 
\STATE $\mathcal{D}_{\rm clip}\leftarrow$ Retain samples from $\mathcal{D}^{\star}$ that $r(s,a)<c$. 
\STATE Update $D$ with $\mathcal{D}_{\rm clip}$ and $\mathcal{D}$.
\STATE Calculate $r(s,a)$ with samples from $\mathcal{D}$.
\STATE Update $\pi$ by a deterministic policy algorithm.
\ENDFOR
\end{algorithmic} 
\end{algorithm}

\begin{table}[htbp]
\centering
\begin{tabular}{l l}
\hline
\textbf{Hyperparameter} & \textbf{Value}\\
\hline
Update frequency &64\\ 
Epoch policy &64\\ 
Batch size &128\\ 
Clipping reward threshold& 5 \\ \hline
\end{tabular}
\caption{Hyperparameters of DDPG-GAIL, TD3-GAIL, SD3-GAIL, and CREDO. }
\label{table_1}
\end{table}

\section{Further Experimental Results}
\renewcommand{\thefigure}{10}
\begin{figure*}[ht]
\centerline{
\begin{minipage}{0.32\linewidth}
\includegraphics[width=\textwidth]{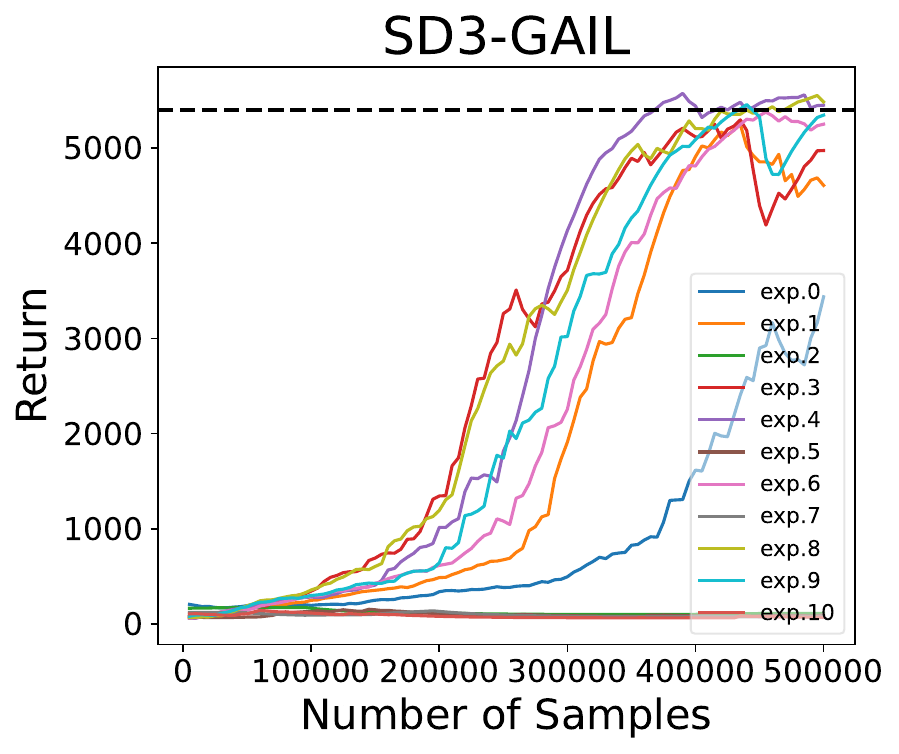}
\end{minipage}
\begin{minipage}{0.32\linewidth}
\includegraphics[width=\textwidth]{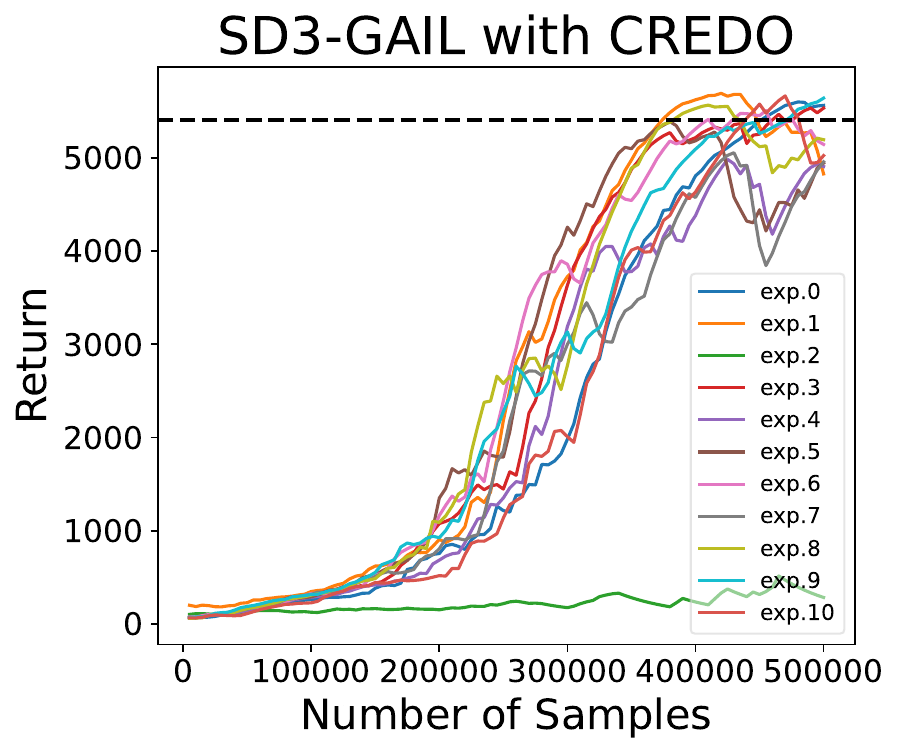}
\end{minipage}
\begin{minipage}{0.32\linewidth}
\includegraphics[width=\textwidth]{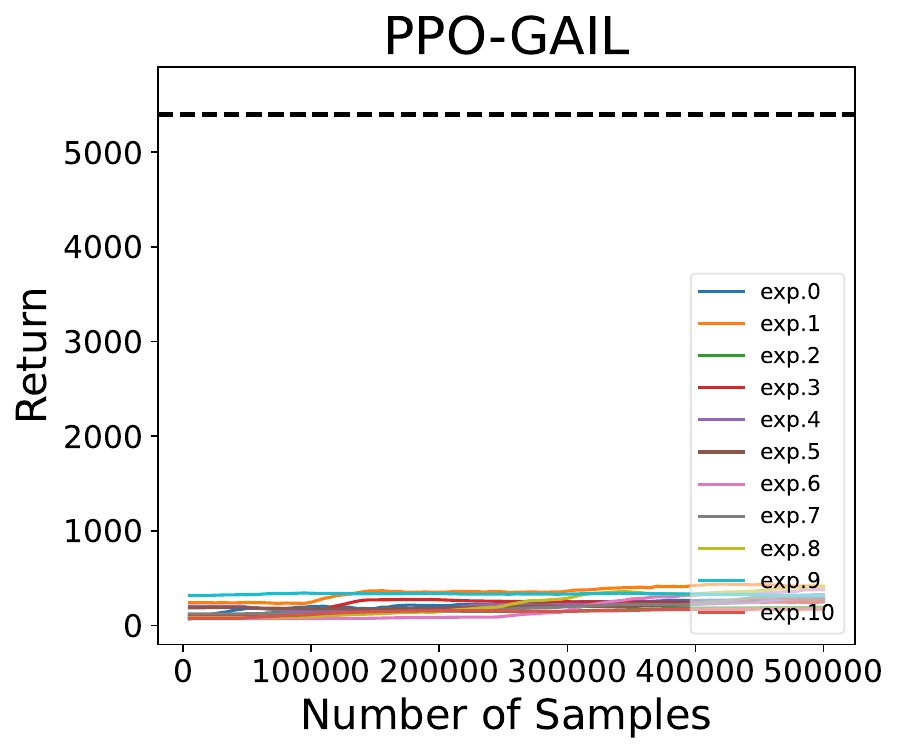}
\end{minipage}}
\caption{Training curves for SD3-GAIL, SD3-GAIL with CREDO and PPO-GAIL in Humanoid-v2. } 
\label{fig:Humanoid}
\end{figure*}

\renewcommand{\thefigure}{11}
\begin{figure*}[ht]
\centerline{
\begin{minipage}{0.32\linewidth}
\includegraphics[width=\textwidth]{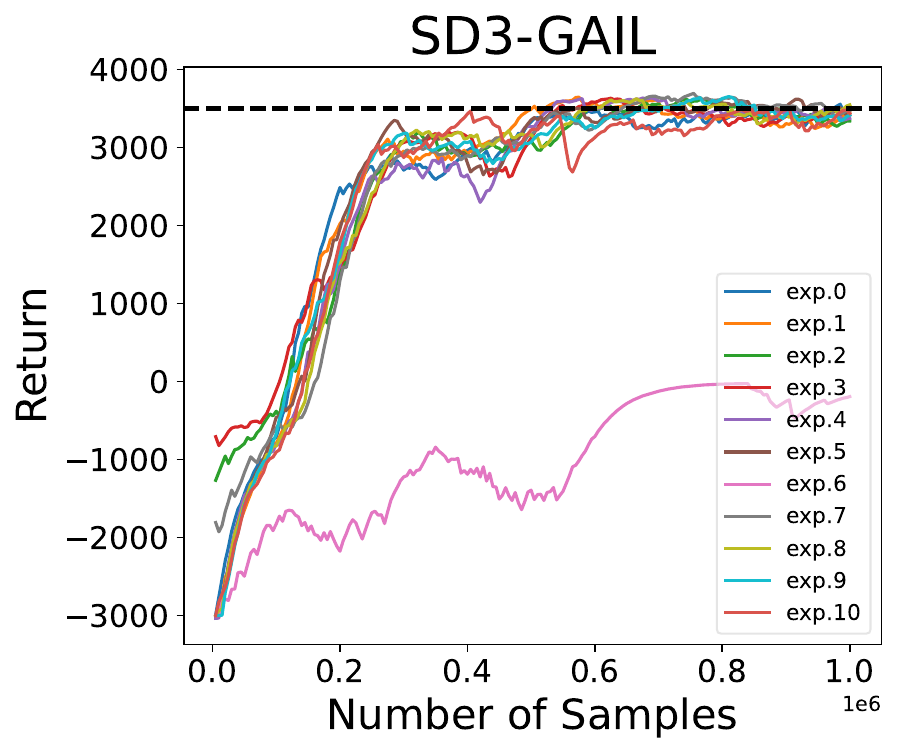}
\end{minipage}
\begin{minipage}{0.32\linewidth}
\includegraphics[width=\textwidth]{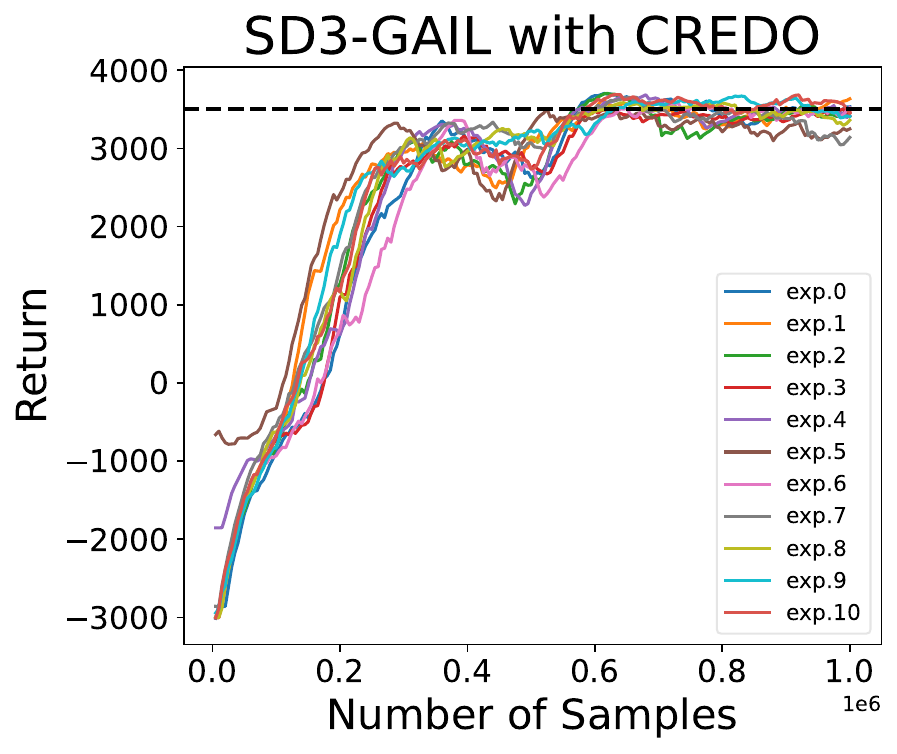}
\end{minipage}
\begin{minipage}{0.32\linewidth}
\includegraphics[width=\textwidth]{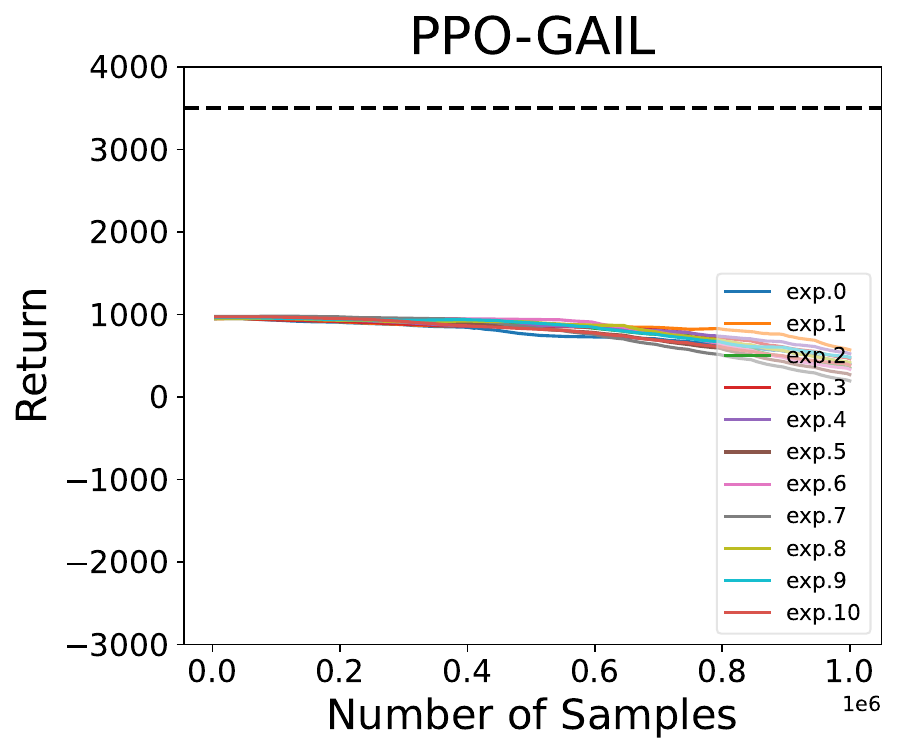}
\end{minipage}}
\caption{Training curves for SD3-GAIL, SD3-GAIL with CREDO and PPO-GAIL in Ant-v2. } 
\label{fig:Ant}
\end{figure*}
\subsection{Hyperparameters}
We train all models on 1 NVIDIA RTX3090 GPU and intel xeon e5-2690 CPU.

Hyperparameters of DDPG-GAIL, TD3-GAIL, SD3-GAIL, and CREDO are kept the same for all experiments. We use an update frequency of 64, batch size of 128, and epoch policy of 128.

\subsection{CREDO}
Fig. \ref{fig:SD3-GAIL-CREDO} shows the return of SD3-GAIL with CREDO among 11 experiments in three MuJoCo environments. It demonstrates the high data efficiency and stable trainability of DE-GAIL with CREDO.

\subsection{TSSG}
Fig. \ref{fig:TSSG_3env} shows that the convergence in TSSG is consistent among different experiments. The absolute gradients and the discriminator value of TSSG are shown in Fig. \ref{fig:TSSG_discriminator} and Fig. \ref{fig:tssg_gradients}, respectively. 

\subsection{Additional Tasks}
Fig. \ref{fig:Humanoid} and Fig. \ref{fig:Ant} depict the learning curves of SD3-GAIL, SD3-GAIL with CREDO and PPO-GAIL in Humanoid-v2 and Ant-v2.

\end{document}